\documentclass[runningheads]{llncs}
\usepackage{graphicx}
\usepackage{amsmath,amssymb} 
\usepackage{color}
\usepackage[utf8]{inputenc}
\usepackage{comment}
\usepackage{gensymb}
\usepackage{pbox}
\usepackage{siunitx}
\usepackage{enumitem}
\usepackage{adjustbox}
\usepackage{subcaption}
\usepackage{tabularx}
\usepackage{booktabs}
\usepackage{colortbl}
\usepackage{tabu}
\usepackage[table,svgnames]{xcolor}
\usepackage{setspace}
\usepackage{textcomp}
\usepackage[skip=1ex,font=small,labelsep=period]{caption}

\definecolor{bblue}{rgb}{0.0,0.25,0.65}
\definecolor{ccol}{rgb}{0.2,0.2,0.2}
\usepackage[pagebackref=true,breaklinks=true,letterpaper=true,colorlinks=true,urlcolor=bblue,bookmarks=false,allcolors=black]{hyperref}

\newcolumntype{Y}{>{\centering\arraybackslash}X}
\newcolumntype{R}{>{\raggedleft\arraybackslash}X}
\newcolumntype{L}{>{\raggedright\arraybackslash}X}

\newenvironment{allintypewriter}{\ttfamily}{\par}
\newcommand{\mytilde}{\raise.17ex\hbox{$\scriptstyle\sim$}}

\newcommand{\eg}{e.g.~}
\newcommand{\ie}{i.e.~}
\newcommand{\etal}{et~al.~}

\newcommand{\wrt}{w.r.t.~}

\newcommand{\mlcellc}[2][c]{%
	\begin{tabular}[#1]{@{}c@{}}#2\end{tabular}}

\begin{document}

\setlength{\abovedisplayskip}{0pt}

\pagestyle{headings}	
\mainmatter
\def\ECCV18SubNumber{1355}

\title{BOP: Benchmark for 6D Object Pose Estimation}

\titlerunning{BOP: Benchmark for 6D Object Pose Estimation}
\authorrunning{Hodaň, Michel et al.}

\newcommand{\namesep}{, }
\author{\small{
 Tomáš~Hodaň$^{1*}$\namesep
 Frank~Michel$^{2*}$\namesep
 Eric~Brachmann$^{3}$\namesep
 Wadim~Kehl$^{4}$
 Anders~Glent~Buch$^{5}$\namesep
 Dirk~Kraft$^{5}$\namesep
 Bertram~Drost$^{6}$\namesep
 Joel~Vidal$^{7}$\namesep
 Stephan~Ihrke$^{2}$
 Xenophon~Zabulis$^{8}$\namesep
 Caner~Sahin$^{9}$\namesep
 Fabian~Manhardt$^{10}$\namesep
 Federico~Tombari$^{10}$
 Tae-Kyun~Kim$^{9}$\namesep
 Jiří~Matas$^{1}$\namesep
 Carsten~Rother$^{3}$
}}
\index{Anders Glent, Buch}
\index{Tae-Kyun, Kim}

\institute{\small{
 $^{1}$CTU~in~Prague\namesep
 $^{2}$TU~Dresden\namesep
 $^{3}$Heidelberg~University\namesep
 $^{4}$Toyota~Research~Institute
 $^{5}$University~of~Southern~Denmark\namesep
 $^{6}$MVTec~Software\namesep
 $^{7}$Taiwan~Tech
 $^{8}$FORTH~Heraklion\namesep
 $^{9}$Imperial~College~London\namesep
 $^{10}$TU~Munich
}}

\maketitle

\begin{abstract}
We propose a benchmark for 6D pose estimation of a rigid object from a single RGB-D input image.
The training data consists of a~texture-mapped 3D object model or images of the object in known 6D poses.
The benchmark comprises of: i)~eight datasets in a unified format that cover different practical scenarios, including two new datasets focusing on varying lighting conditions, ii)~an evaluation methodology with a pose-error function that deals with pose ambiguities, iii)~a comprehensive evaluation of 15 diverse recent methods that captures the status quo of the field, and iv)~an online evaluation system that is open for continuous submission of new results. The evaluation shows that methods based on point-pair features currently perform best, outperforming template matching methods, learning-based methods and methods based on 3D local features.
The project website is available at \texttt{\href{http://bop.felk.cvut.cz}{bop.felk.cvut.cz}}. \let\thefootnote\relax\footnotetext{$^{*}$Authors have been leading the project jointly.}
\end{abstract}

\section{Introduction}

Estimating the 6D pose, \ie 3D translation and 3D rotation, of a rigid object has become an accessible task with the introduction of consumer-grade RGB-D sensors. An accurate, fast and robust method that solves this task will have a big impact in application fields such as robotics or augmented reality.

\begin{figure}[!t]
	\begin{center}
		\begin{allintypewriter}

		\setlength{\tabcolsep}{1pt} 
		\renewcommand{\arraystretch}{1} 

		\begingroup
		\setlength{\tabcolsep}{0pt} 
		\renewcommand{\arraystretch}{0.6} 
		\begin{tabularx}{\textwidth}{ *{7}{Y} }	
			{\tiny \mlcellc[b]{LM/LM-O \cite{hinterstoisser2012accv,brachmann2014learning}}} &
			{\tiny \mlcellc[b]{IC-MI} \cite{tejani2014latent}} &
			{\tiny \mlcellc[b]{IC-BIN} \cite{doumanoglou2016recovering}} &
			{\tiny \mlcellc[b]{T-LESS \cite{hodan2017tless}}} &
			{\tiny \mlcellc[b]{RU-APC \cite{rennie2016dataset}}} &
			{\tiny \mlcellc[b]{TUD-L - new}} &
			{\tiny \mlcellc[b]{TYO-L - new}} \\
			\includegraphics[width=0.137\columnwidth]{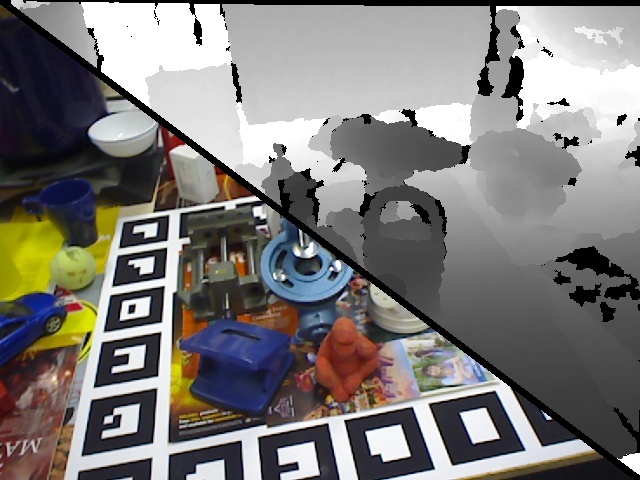} &
			\includegraphics[width=0.137\columnwidth]{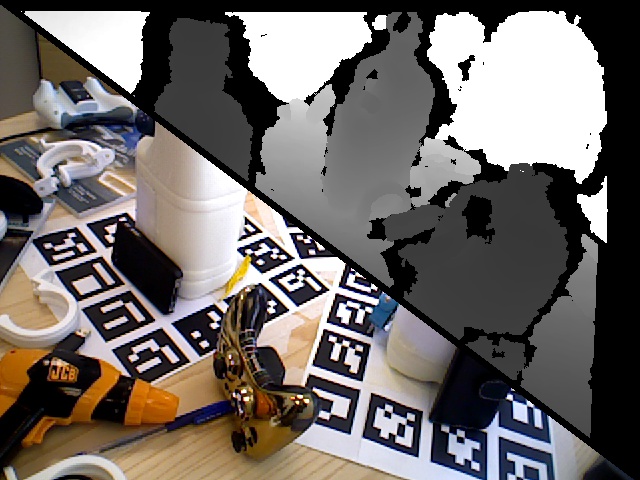} &
			\includegraphics[width=0.137\columnwidth]{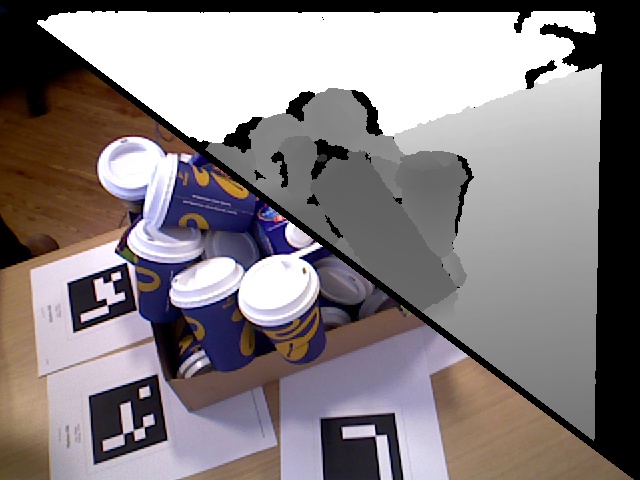} &
			\includegraphics[width=0.137\columnwidth]{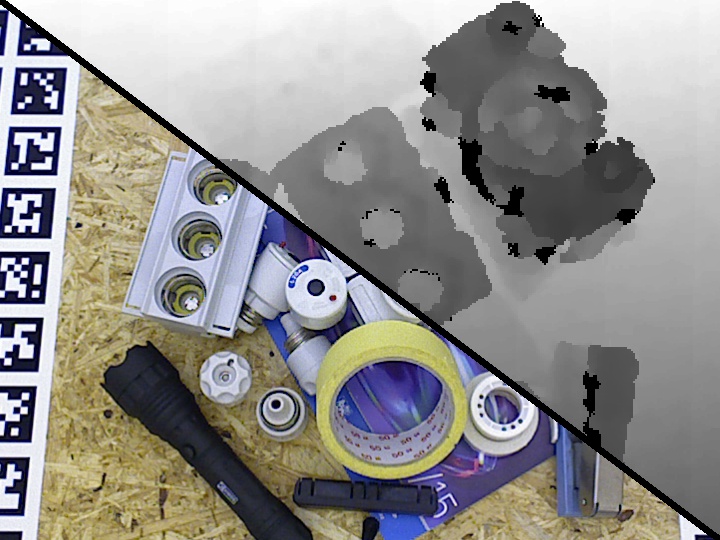} &
			\includegraphics[width=0.137\columnwidth]{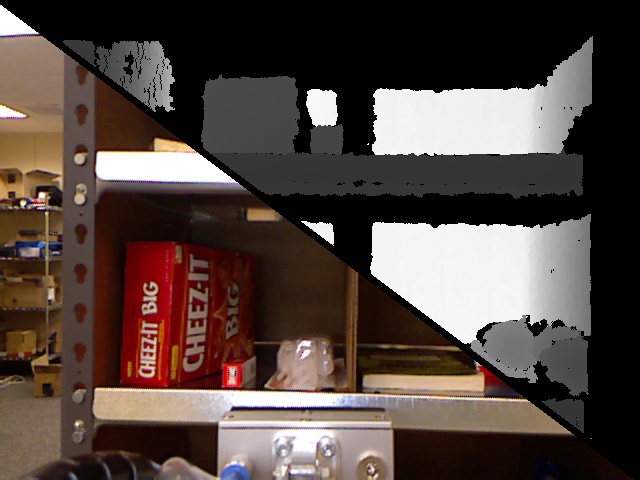} &
			\includegraphics[width=0.137\columnwidth]{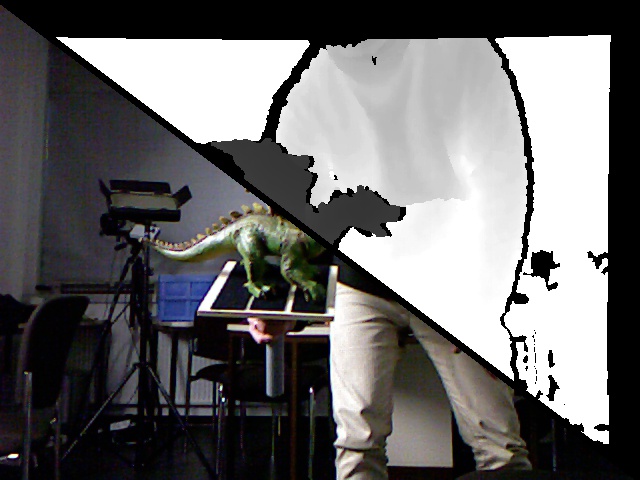} &
			\includegraphics[width=0.137\columnwidth]{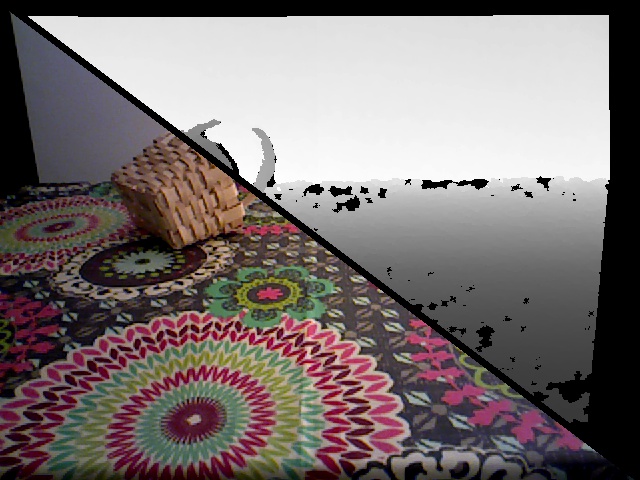} \\
			
			\includegraphics[width=0.137\columnwidth]{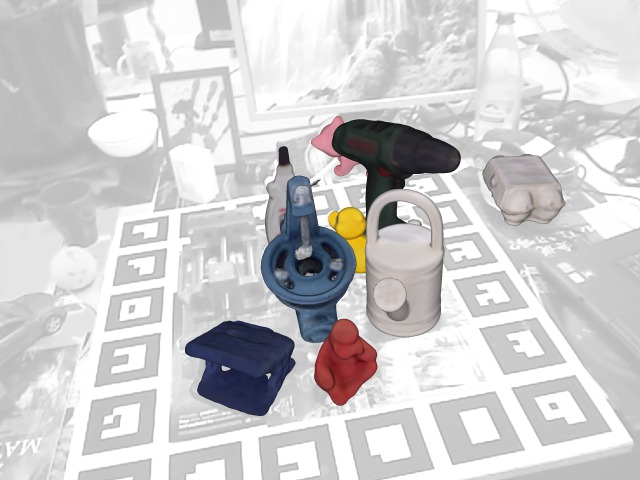} &
			\includegraphics[width=0.137\columnwidth]{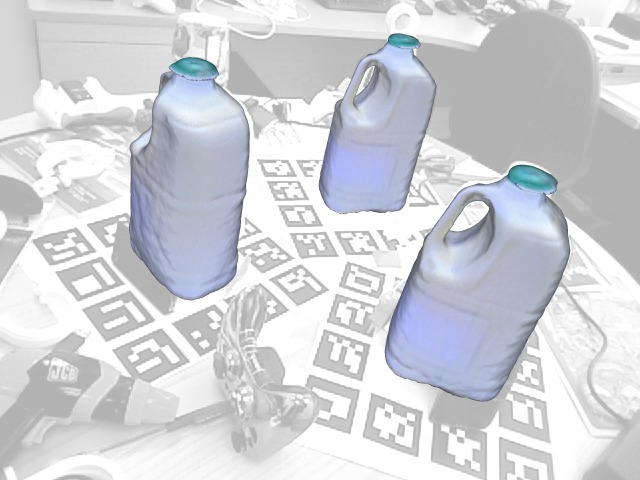} &
			\includegraphics[width=0.137\columnwidth]{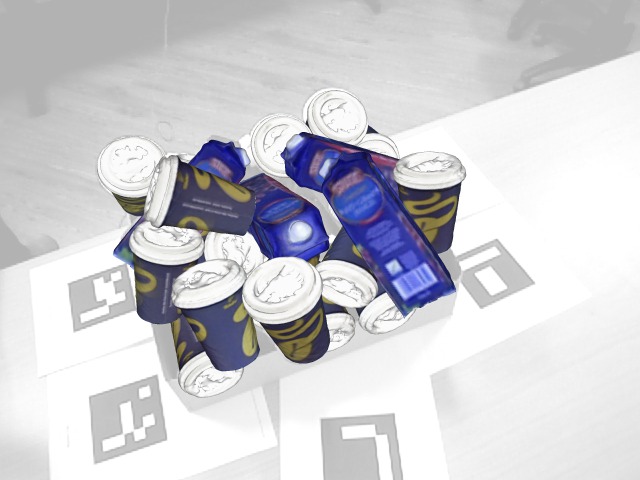} &
			\includegraphics[width=0.137\columnwidth]{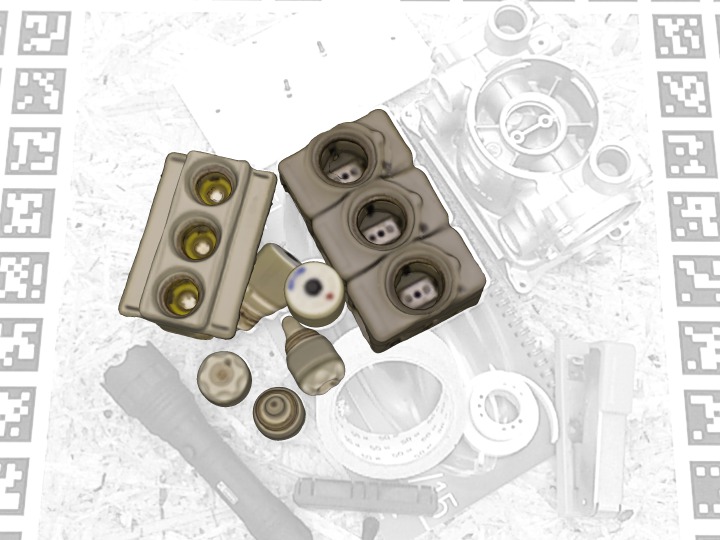} &
			\includegraphics[width=0.137\columnwidth]{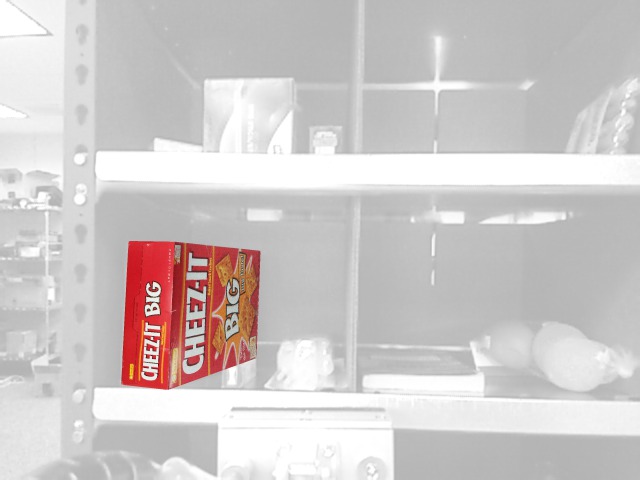} &
			\includegraphics[width=0.137\columnwidth]{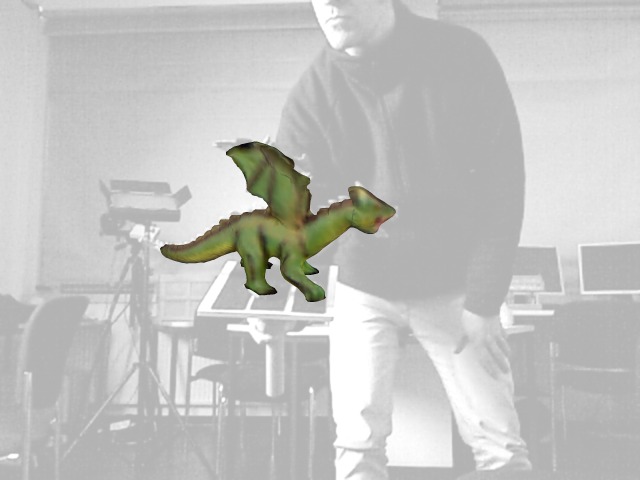} &
			\includegraphics[width=0.137\columnwidth]{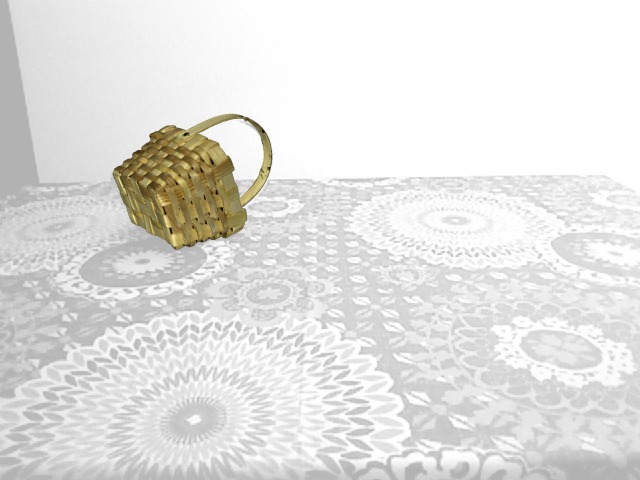}
		\end{tabularx}
		\endgroup

		\vspace{0.5ex}

		{\tiny
		\begin{tabu}{*{30}{c}}
		\includegraphics[scale=0.161]{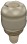} &
		\includegraphics[scale=0.161]{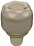} &
		\includegraphics[scale=0.161]{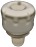} &
		\includegraphics[scale=0.161]{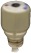} &
		\includegraphics[scale=0.161]{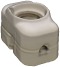} &
		\includegraphics[scale=0.161]{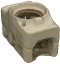} &
		\includegraphics[scale=0.161]{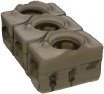} &
		\includegraphics[scale=0.161]{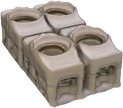} &
		\includegraphics[scale=0.161]{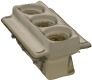} &
		\includegraphics[scale=0.161]{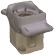} &
		\includegraphics[scale=0.161]{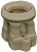} &
		\includegraphics[scale=0.161]{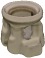} &
		\includegraphics[scale=0.161]{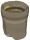} &
		\includegraphics[scale=0.161]{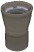} &
		\includegraphics[scale=0.161]{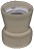} &
		\includegraphics[scale=0.161]{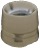} &
		\includegraphics[scale=0.161]{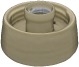} &
		\includegraphics[scale=0.161]{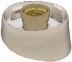} &
		\includegraphics[scale=0.161]{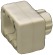} &
		\includegraphics[scale=0.161]{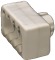} &
		\includegraphics[scale=0.161]{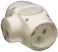} &
		\includegraphics[scale=0.161]{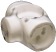} &
		\includegraphics[scale=0.161]{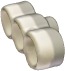} &
		\includegraphics[scale=0.161]{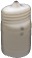} &
		\includegraphics[scale=0.161]{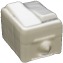} &
		\includegraphics[scale=0.161]{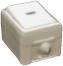} &
		\includegraphics[scale=0.161]{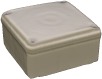} &
		\includegraphics[scale=0.161]{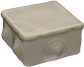} &
		\includegraphics[scale=0.161]{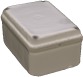} &
		\includegraphics[scale=0.161]{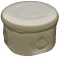}
		\\

		\rowfont{\color{black!75}}1 & 2 & 3 & 4 & 5 & 6 & 7 & 8 & 9 & 10 & 11 & 12 & 13 & 14 & 15 & 16 & 17 & 18 & 19 & 20 & 21 & 22 & 23 & 24 & 25 & 26 & 27 & 28 & 29 & 30
		\\ \multicolumn{30}{c}{T-LESS}
		\end{tabu} \\

		\begin{tabu}{*{22}{c}}
		\includegraphics[scale=0.161]{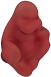} &
		\includegraphics[scale=0.161]{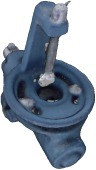} &
		\includegraphics[scale=0.161]{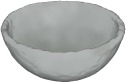} &
		\includegraphics[scale=0.161]{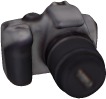} &
		\includegraphics[scale=0.161]{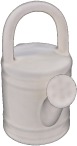} &
		\includegraphics[scale=0.161]{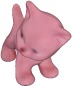} &
		\includegraphics[scale=0.161]{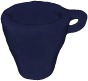} &
		\includegraphics[scale=0.161]{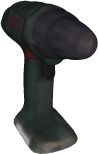} &
		\includegraphics[scale=0.161]{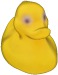} &
		\includegraphics[scale=0.161]{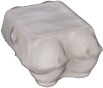} &
		\includegraphics[scale=0.161]{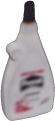} &
		\includegraphics[scale=0.161]{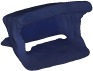} &
		\includegraphics[scale=0.161]{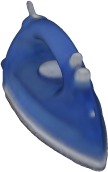} &
		\includegraphics[scale=0.161]{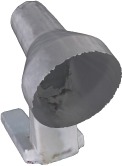} &
		\includegraphics[scale=0.161]{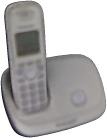} &
		&
		\includegraphics[scale=0.161]{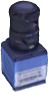} &
		\includegraphics[scale=0.161]{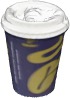} &
		\includegraphics[scale=0.161]{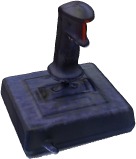} &
		\includegraphics[scale=0.161]{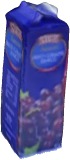} &
		\includegraphics[scale=0.161]{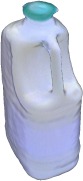} &
		\includegraphics[scale=0.161]{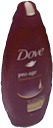}
		\\

		\rowfont{\color{black!75}}1 & 2 & 3 & 4 & 5 & 6 & 7 & 8 & 9 & 10 & 11 & 12 & 13 & 14 & 15 & \hspace{2em} & 1 & 2 & 3 & 4 & 5 & 6
		\\ \multicolumn{15}{c}{LM/LM-O} & & \multicolumn{6}{c}{IC-MI/IC-BIN}
		\end{tabu} \\

		\begin{tabu}{*{21}{c}}
		\includegraphics[scale=0.161]{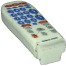} &
		\includegraphics[scale=0.161]{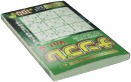} &
		\includegraphics[scale=0.161]{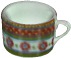} &
		\includegraphics[scale=0.161]{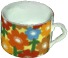} &
		\includegraphics[scale=0.161]{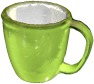} &
		\includegraphics[scale=0.161]{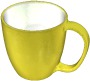} &
		\includegraphics[scale=0.161]{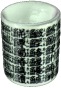} &
		\includegraphics[scale=0.161]{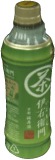} &
		\includegraphics[scale=0.161]{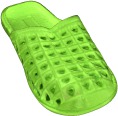} &
		\includegraphics[scale=0.161]{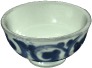} &
		\includegraphics[scale=0.161]{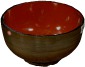} &
		\includegraphics[scale=0.161]{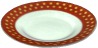} &
		\includegraphics[scale=0.161]{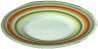} &
		\includegraphics[scale=0.161]{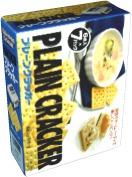} &
		\includegraphics[scale=0.161]{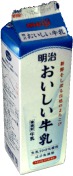} &
		\includegraphics[scale=0.161]{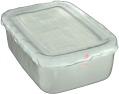} &
		\includegraphics[scale=0.161]{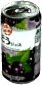} &
		\includegraphics[scale=0.161]{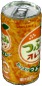} &
		\includegraphics[scale=0.161]{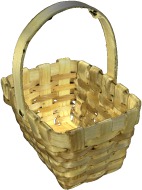} &
		\includegraphics[scale=0.161]{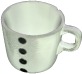} &
		\includegraphics[scale=0.161]{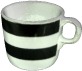}
		\\
		\rowfont{\color{black!75}}1 & 2 & 3 & 4 & 5 & 6 & 7 & 8 & 9 & 10 & 11 & 12 & 13 & 14 & 15 & 16 & 17 & 18 & 19 & 20 & 21
		\\ \multicolumn{21}{c}{TYO-L}
		\end{tabu} \\

		\begin{tabu}{*{18}{c}}
		\includegraphics[scale=0.161]{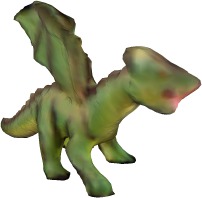} &
		\includegraphics[scale=0.161]{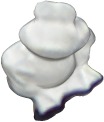} &
		\includegraphics[scale=0.161]{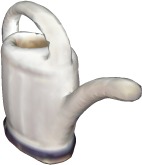} &
		&
		\includegraphics[scale=0.161]{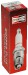} &
		\includegraphics[scale=0.161]{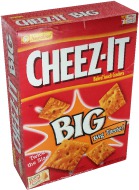} &
		\includegraphics[scale=0.161]{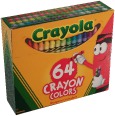} &
		\includegraphics[scale=0.161]{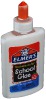} &
		\includegraphics[scale=0.161]{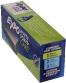} &
		\includegraphics[scale=0.161]{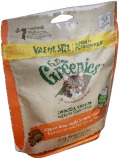} &
		\includegraphics[scale=0.161]{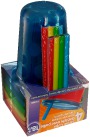} &
		\includegraphics[scale=0.161]{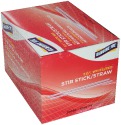} &
		\includegraphics[scale=0.161]{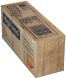} &
		\includegraphics[scale=0.161]{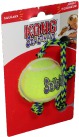} &
		\includegraphics[scale=0.161]{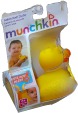} &
		\includegraphics[scale=0.161]{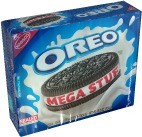} &
		\includegraphics[scale=0.161]{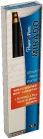} &
		\includegraphics[scale=0.161]{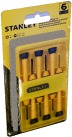}
		\\
		\rowfont{\color{black!75}}1 & 2 & 3 & \hspace{2em} & 1 & 2 & 3 & 4 & 5 & 6 & 7 & 8 & 9 & 10 & 11 & 12 & 13 & 14
		\\ \multicolumn{3}{c}{TUD-L} & & \multicolumn{14}{c}{RU-APC}
		\end{tabu}
		}

		\vspace{0.5ex}
		
		\caption{\label{fig:datasets_overview} A collection of benchmark datasets. Top: Example test RGB-D images where the second row shows the images overlaid with 3D object models in the ground-truth 6D poses. Bottom: Texture-mapped 3D object models. At training time, a method is given an object model or a set of training images with ground-truth object poses. At test time, the method is provided with one test image and an identifier of the target object. The task is to estimate the 6D pose of an instance of this object.}
	\end{allintypewriter}
	\end{center}
\end{figure}

Many methods for 6D object pose estimation have been published recently, \eg \cite{tejani2014latent,krull2015learning,hodan2015detection,brachmann2016uncertainty,wohlhart2015learning,kehl2017ssd,rad2017bb8,michel2017global}, but it is unclear which methods perform well and in which scenarios.
The most commonly used dataset for evaluation was created by Hinterstoisser \etal \cite{hinterstoisser2012accv}, which was not intended as a general benchmark and has several limitations: the lighting conditions are constant and the objects are easy to distinguish, unoccluded and located around the image center. Since then, some of the limitations have been addressed. Brachmann \etal \cite{brachmann2014learning} added ground-truth annotation for occluded objects in the dataset of \cite{hinterstoisser2012accv}. Hodaň~\etal \cite{hodan2017tless} created a dataset that features industry-relevant objects with symmetries and similarities, and Drost \etal \cite{drost2017introducing} introduced a dataset containing objects with reflective surfaces. However, the datasets have different formats and no standard evaluation methodology has emerged. New methods are usually compared with only a few competitors on a small subset of datasets. \pagebreak

\noindent This work makes the following contributions:

\begin{enumerate}[leftmargin=3.4ex]
\itemsep0em

\item \textbf{Eight datasets in a unified format}, including
two new datasets focusing on varying lighting conditions, are made available (Fig.~\ref{fig:datasets_overview}).
The datasets~contain: i)~texture-mapped 3D models of 89~objects with a wide range of sizes, shapes and reflectance properties, ii)~277K training RGB-D images showing isolated objects from different viewpoints, and iii)~62K test RGB-D images of scenes with graded complexity. High-quality ground-truth 6D~poses of the modeled objects are provided for all images.

\item \textbf{An evaluation methodology} based on~\cite{hodan2016evaluation} that includes the formulation of an industry-relevant task, and a pose-error function which deals well with pose ambiguity of symmetric or partially occluded objects, in contrast to the commonly used function by Hinterstoisser \etal\cite{hinterstoisser2012accv}.

\item \textbf{A comprehensive evaluation} of 15 methods on the benchmark datasets using the proposed evaluation methodology. We provide an analysis of the results, report the state of the art, and identify open problems.

\item \textbf{An online evaluation system} at \texttt{\href{http://bop.felk.cvut.cz}{bop.felk.cvut.cz}} that allows for continuous submission of new results and provides up-to-date leaderboards.

\end{enumerate} \pagebreak

\subsection{Related Work}

The progress of research in computer vision has been strongly influenced by challenges and benchmarks, which enable to evaluate and compare methods and better understand their limitations. The Middlebury benchmark~\cite{scharstein2002taxonomy,scharstein2007learning} for depth from stereo and optical flow estimation was one of the first that gained large attention. The PASCAL VOC challenge~\cite{everingham2010pascal}, based on a photo collection from the internet, was the first to standardize the evaluation of object detection and image classification. It was followed by 
the ImageNet challenge~\cite{russakovsky2015imagenet}, which has been running for eight years, starting in 2010, and has pushed image classification methods to new levels of accuracy. The key was a large-scale dataset that enabled training of deep neural networks, which then quickly became a game-changer for many other tasks~\cite{krizhevsky2012imagenet}.
With increasing maturity of computer vision methods, recent benchmarks moved to real-world scenarios. A great example is the KITTI benchmark \cite{geiger2012kitti} focusing on problems related to autonomous driving. It showed that methods ranking high on established benchmarks, such as the Middlebury, perform below average when moved outside the laboratory conditions.

Unlike the PASCAL VOC and ImageNet challenges, the task considered in this work requires a specific set of calibrated modalities that cannot be easily acquired from the internet.
In contrast to KITTY, it was not necessary to record large amounts of new data. By combining existing datasets, we have
covered many practical scenarios. Additionally, we created two datasets with varying lighting conditions, which is an aspect not covered by the existing datasets.

\section{Evaluation Methodology} \label{sec:methodology}

The proposed evaluation methodology formulates the 6D object pose estimation
task and defines a pose-error function which is compared with the commonly used function by Hinterstoisser \etal\cite{hinterstoisser2012pami}.

\subsection{Formulation of the Task} \label{sec:task_formulation}

Methods for 6D object pose estimation report their predictions on the basis of two sources of information. Firstly, at training time, a method is given a training set $T = \{T_o\}_{o=1}^n$, where $o$ is an object identifier. Training data $T_o$ may have different forms,
\eg a 3D mesh model of the object or a set of
RGB-D images showing  object instances in known 6D poses. Secondly, at test time, the method is provided with a test target defined by a pair $(I, o)$, where $I$ is an image showing at least one instance of object $o$.
The goal is to estimate the 6D pose of one of the instances of object $o$ visible in image $I$.

If multiple instances of the same object model are present, then the pose of an arbitrary instance
may be reported.
If multiple object models are shown in a test image, and annotated with their ground truth poses, then each object model may define a different test target. 
For example, if a test image shows three object models, each in two instances, then we define three test targets. For each test target, the pose of one of the two object instances has to be estimated.

This task reflects the industry-relevant bin-picking scenario where a robot needs to grasp a single arbitrary instance of the required object, \eg a~component such as a bolt or nut, and perform some operation with it.
It is the simplest variant of the 6D localization task~\cite{hodan2016evaluation} and a common denominator of its other variants, which deal with a single instance of multiple objects, multiple instances of a single object, or multiple instances of multiple objects. It is also the core of the 6D detection task, where no prior information about the object presence in the test image is provided~\cite{hodan2016evaluation}.

\subsection{Measuring Error} \label{sec:error}

A 3D object model is defined as a set of vertices in $\mathbb{R}^3$ and a set of polygons that describe the object surface.
The object pose is represented by a $4\times4$ matrix $\mathbf{P} = [\mathbf{R} , \mathbf{t}; \mathbf{0}, 1]$, where $\mathbf{R}$ is a $3\times3$ rotation matrix and $\mathbf{t}$ is a $3\times1$ translation vector. The matrix $\mathbf{P}$ transforms a 3D homogeneous point $\mathbf{x}_m$ in the model coordinate system to a 3D point $\mathbf{x}_c$ in the camera coordinate system:~$\mathbf{x}_c = \mathbf{P}\mathbf{x}_m$.

\subsubsection{Visible Surface Discrepancy.}
To calculate the error of an estimated pose~$\hat{\mathbf{P}}$ \wrt the ground-truth pose~$\bar{\mathbf{P}}$ in a test image $I$, an object model~$\mathcal{M}$ is first rendered in the two poses.
The result of the rendering is two distance maps\footnote{A distance map stores at a pixel~$p$ the distance from the camera center to a 3D point $\mathbf{x}_p$ that projects to $p$. It can be readily computed from the depth map which stores at $p$ the $Z$ coordinate of $\mathbf{x}_p$ and which can be obtained by a Kinect-like sensor.} $\hat{S}$ and~$\bar{S}$. As in \cite{hodan2016evaluation}, the distance maps are compared with the distance map $S_I$ of the test image~$I$ to obtain the visibility masks $\hat{V}$ and $\bar{V}$, \ie the sets of pixels where the model $\mathcal{M}$ is visible in the image $I$ (Fig.~\ref{fig:vsd_components}).
Given a misalignment tolerance $\tau$, the error is calculated as:

\begin{equation} \label{eq:vsd}
e_\mathrm{VSD}(\hat{S}, \bar{S}, S_I, \hat{V}, \bar{V}, \tau) =
\underset{p \in \hat{V} \cup \bar{V}}{\mathrm{avg}}
\begin{cases} 
0 & \text{if $p \in \hat{V} \cap \bar{V} \, \wedge \, |\hat{S}(p) - \bar{S}(p)| < \tau$} \\
1 & \text{otherwise}.
\end{cases}
\end{equation}

\def\pboxc{\relax\ifvmode\centering\fi}

\begin{figure}[!t]
    \begin{center}

        \begingroup
        \setlength{\tabcolsep}{1.5pt} 
        \renewcommand{\arraystretch}{0.9} 

        \begin{tabular}{ c c c c c c c c }
        	\pbox{\textwidth}{\pboxc{}\scriptsize{$RGB_I$} \\ \vspace{0.5ex}} &
        	\pbox{\textwidth}{\pboxc{}\scriptsize{$S_I$} \\ \vspace{0.5ex}} &
        	\hspace{1em} &
        	\pbox{\textwidth}{\pboxc{}\scriptsize{$\hat{S}$} \\ \vspace{0.5ex}} &
        	\pbox{\textwidth}{\pboxc{}\scriptsize{$\hat{V}$} \\ \vspace{0.5ex}} &
        	\pbox{\textwidth}{\pboxc{}\scriptsize{$\bar{S}$} \\ \vspace{0.5ex}} &
        	\pbox{\textwidth}{\pboxc{}\scriptsize{$\bar{V}$} \\ \vspace{0.5ex}} &
        	\pbox{\textwidth}{\pboxc{}\scriptsize{$S_\mathrm{\Delta}$} \\ \vspace{0.5ex}} \\
        	
	        \includegraphics[width=0.117\columnwidth]{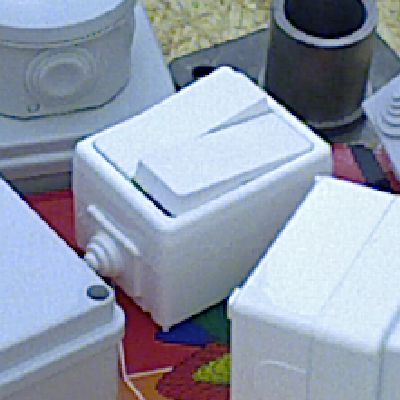} &
        	\includegraphics[width=0.117\columnwidth]{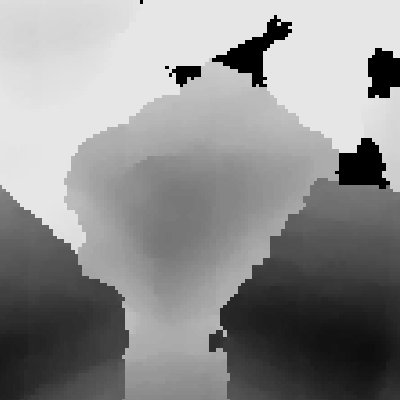} &
            &
        	\includegraphics[width=0.117\columnwidth]{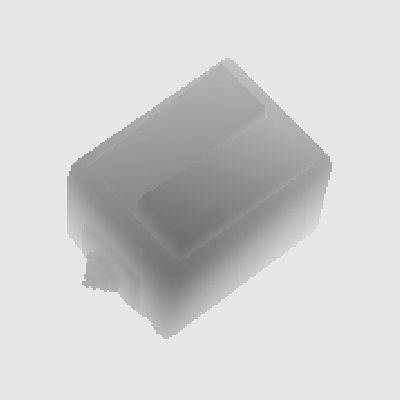} &
        	\includegraphics[width=0.117\columnwidth]{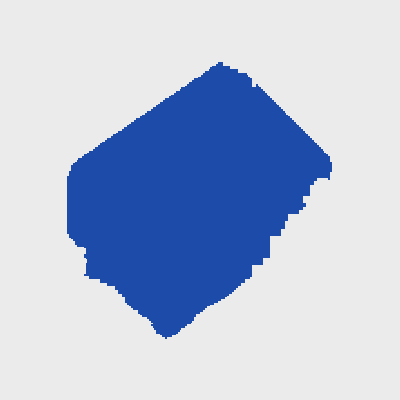} &
        	\includegraphics[width=0.117\columnwidth]{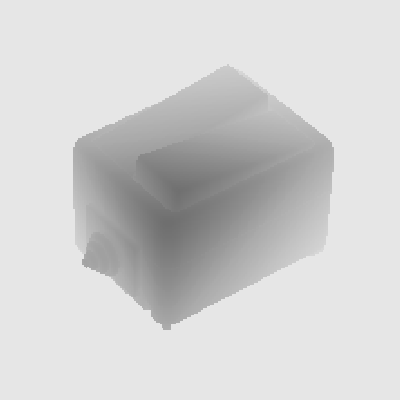} &
        	\includegraphics[width=0.117\columnwidth]{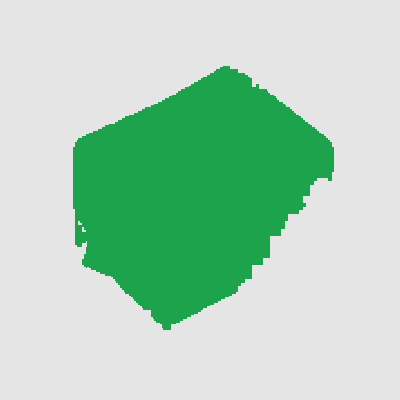} &
        	\includegraphics[width=0.117\columnwidth]{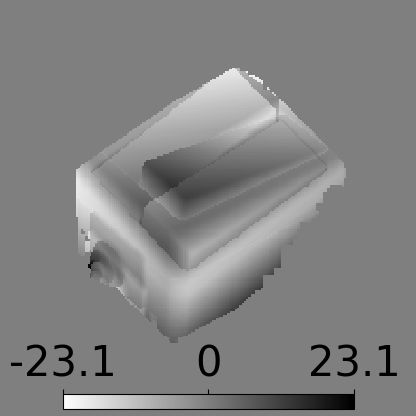} \\
        \end{tabular}
        \caption{Quantities used in the calculation of $e_\mathrm{VSD}$.
        Left: Color channels $RGB_I$ (only for illustration) and distance map $S_I$ of a test image~$I$.
        Right: Distance maps $\hat{S}$ and $\bar{S}$ are obtained by rendering the object model $\mathcal{M}$ at the estimated pose $\hat{\mathbf{P}}$ and the ground-truth pose $\bar{\mathbf{P}}$ respectively. $\hat{V}$ and $\bar{V}$ are masks of the model surface that is visible in~$I$, obtained by comparing $\hat{S}$ and $\bar{S}$ with $S_I$. Distance differences $S_\mathrm{\Delta}(p) = \hat{S}(p) - \bar{S}(p)$, $\forall p \in \hat{V} \cap \bar{V}$, are used for the pixel-wise evaluation of the surface alignment.}
        \label{fig:vsd_components}

        \vspace{6.5ex}

        \begin{tabular}{ c c c c c c c c }
            \pbox{\textwidth}{\pboxc{}\scriptsize{a: \textbf{0.04}} \\ \scriptsize{3.7/15.2} \\ \vspace{0.5ex}} &
            \pbox{\textwidth}{\pboxc{}\scriptsize{b: \textbf{0.08}} \\ \scriptsize{3.6/10.9} \\ \vspace{0.5ex}} &
            \pbox{\textwidth}{\pboxc{}\scriptsize{c: \textbf{0.11}} \\ \scriptsize{3.2/13.4} \\ \vspace{0.5ex}} &
            \pbox{\textwidth}{\pboxc{}\scriptsize{d: \textbf{0.19}} \\ \scriptsize{1.0/6.4} \\ \vspace{0.5ex}} &
            \pbox{\textwidth}{\pboxc{}\scriptsize{e: \textbf{0.28}} \\ \scriptsize{1.4/7.7} \\ \vspace{0.5ex}} &
            \pbox{\textwidth}{\pboxc{}\scriptsize{f: \textbf{0.34}} \\ \scriptsize{2.1/6.4} \\ \vspace{0.5ex}} &
            \pbox{\textwidth}{\pboxc{}\scriptsize{g: \textbf{0.40}} \\ \scriptsize{2.1/8.6} \\ \vspace{0.5ex}} &
            \pbox{\textwidth}{\pboxc{}\scriptsize{h: \textbf{0.44}} \\ \scriptsize{4.8/21.7} \\ \vspace{0.5ex}} \\
            
            \includegraphics[width=0.117\columnwidth]{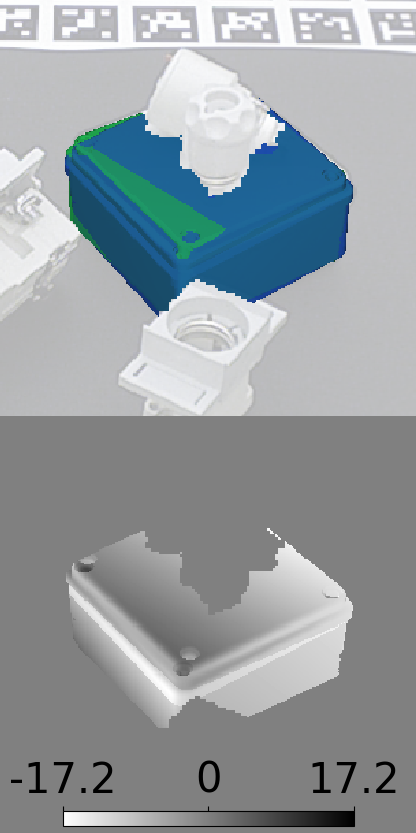} &
            \includegraphics[width=0.117\columnwidth]{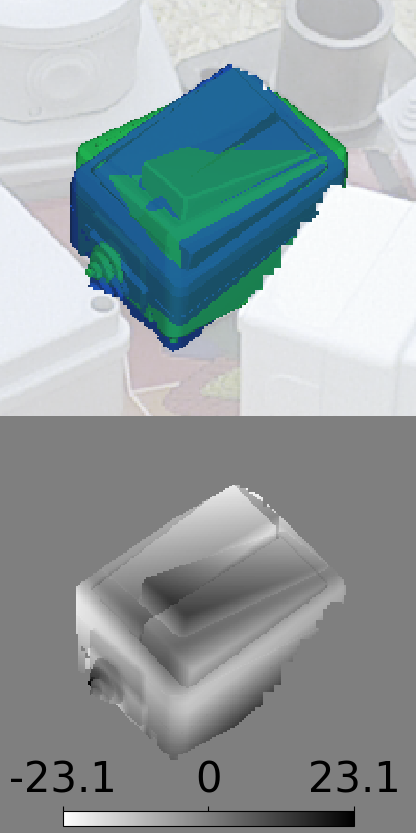} &
            \includegraphics[width=0.117\columnwidth]{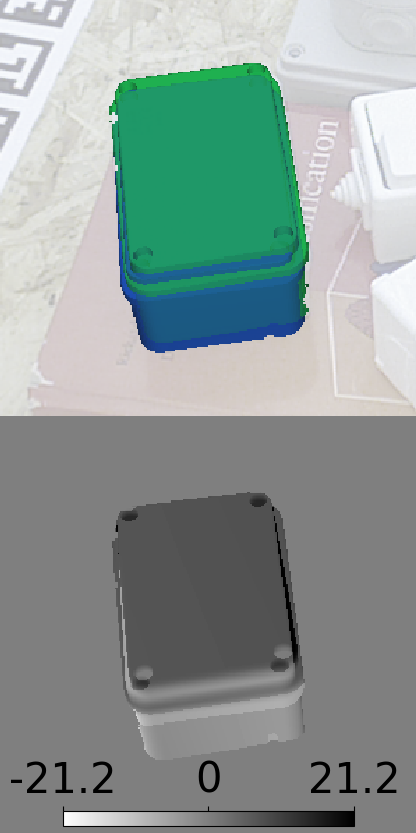} &
            \includegraphics[width=0.117\columnwidth]{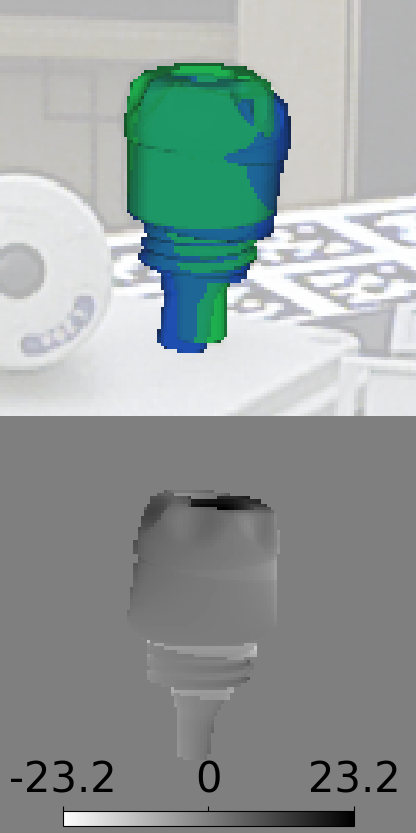} &
            \includegraphics[width=0.117\columnwidth]{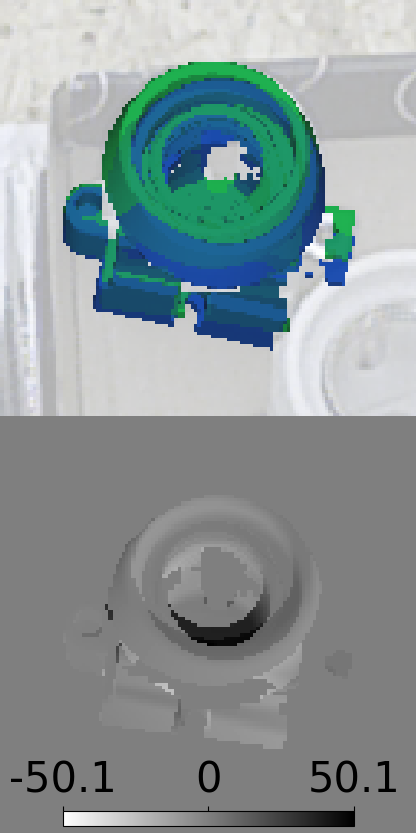} &
            \includegraphics[width=0.117\columnwidth]{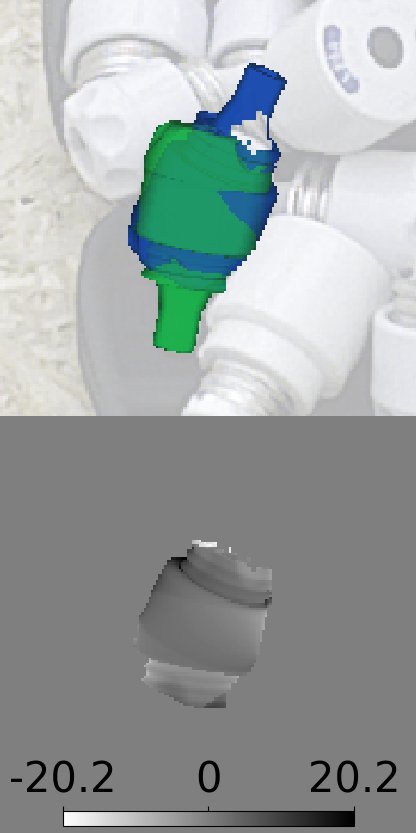} &
            \includegraphics[width=0.117\columnwidth]{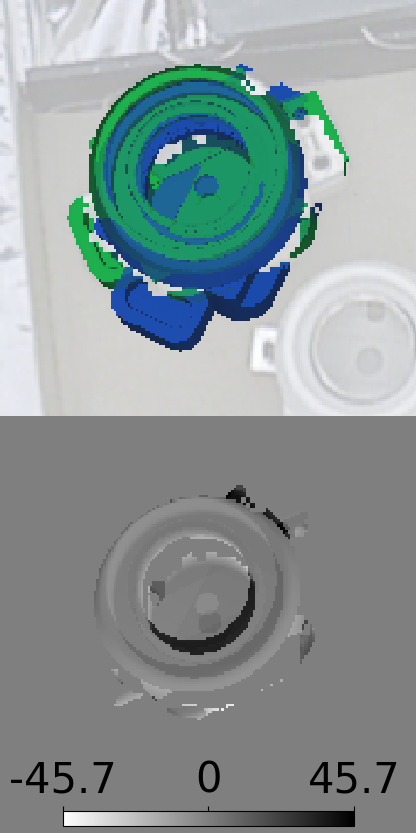} &
            \includegraphics[width=0.117\columnwidth]{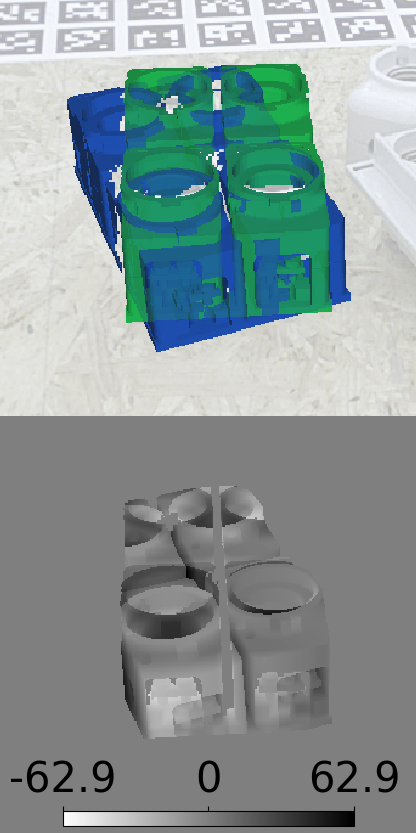} \\
            
            \pbox{\textwidth}{\pboxc{}\vspace{0.7ex}\scriptsize{i: \textbf{0.47}} \\ \scriptsize{4.8/9.2} \\ \vspace{0.5ex}} &
            \pbox{\textwidth}{\pboxc{}\vspace{0.7ex}\scriptsize{j: \textbf{0.54}} \\ \scriptsize{6.9/10.8} \\ \vspace{0.5ex}} &
            \pbox{\textwidth}{\pboxc{}\vspace{0.7ex}\scriptsize{k: \textbf{0.57}} \\ \scriptsize{6.9/8.9} \\ \vspace{0.5ex}} &
            \pbox{\textwidth}{\pboxc{}\vspace{0.7ex}\scriptsize{l: \textbf{0.64}} \\ \scriptsize{21.0/21.7} \\ \vspace{0.5ex}} &
            \pbox{\textwidth}{\pboxc{}\vspace{0.7ex}\scriptsize{m: \textbf{0.66}} \\ \scriptsize{4.4/6.5} \\ \vspace{0.5ex}} &
            \pbox{\textwidth}{\pboxc{}\vspace{0.7ex}\scriptsize{n: \textbf{0.76}} \\ \scriptsize{8.8/9.9} \\ \vspace{0.5ex}} &
            \pbox{\textwidth}{\pboxc{}\vspace{0.7ex}\scriptsize{o: \textbf{0.89}} \\ \scriptsize{49.4/11.1} \\ \vspace{0.5ex}} &
            \pbox{\textwidth}{\pboxc{}\vspace{0.7ex}\scriptsize{p: \textbf{0.95}} \\ \scriptsize{32.8/10.8} \\ \vspace{0.5ex}} \\
            
            \includegraphics[width=0.117\columnwidth]{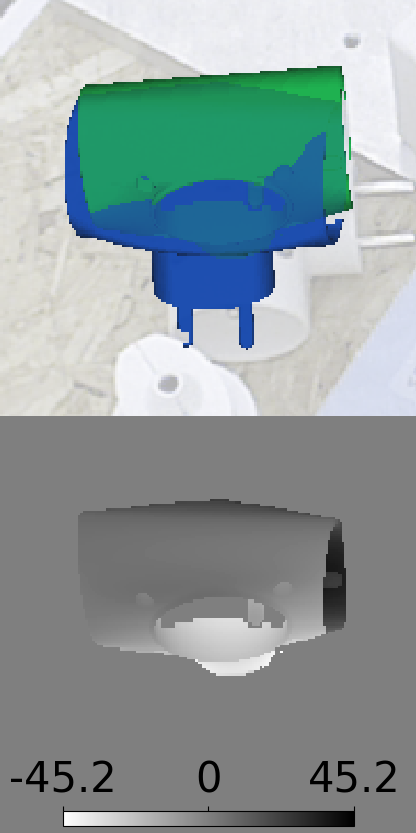} &
            \includegraphics[width=0.117\columnwidth]{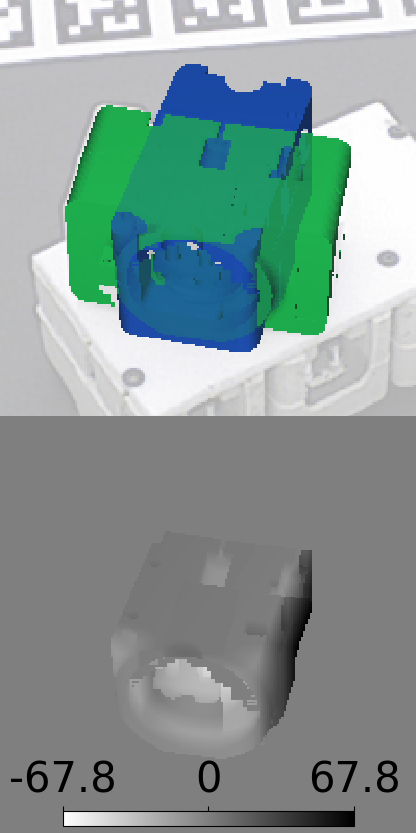} &
            \includegraphics[width=0.117\columnwidth]{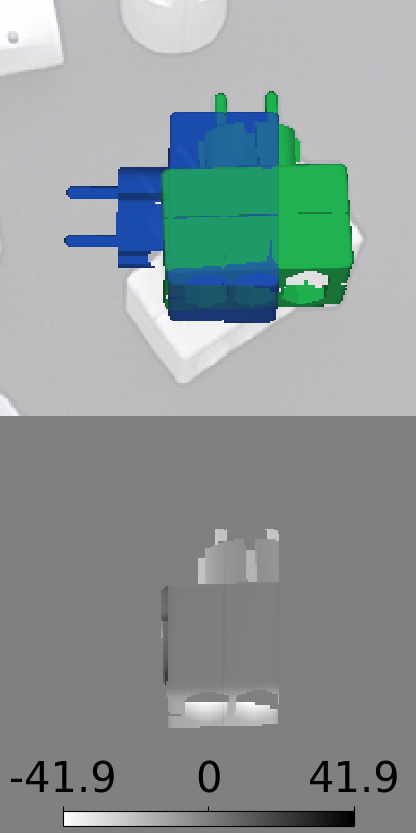} &
            \includegraphics[width=0.117\columnwidth]{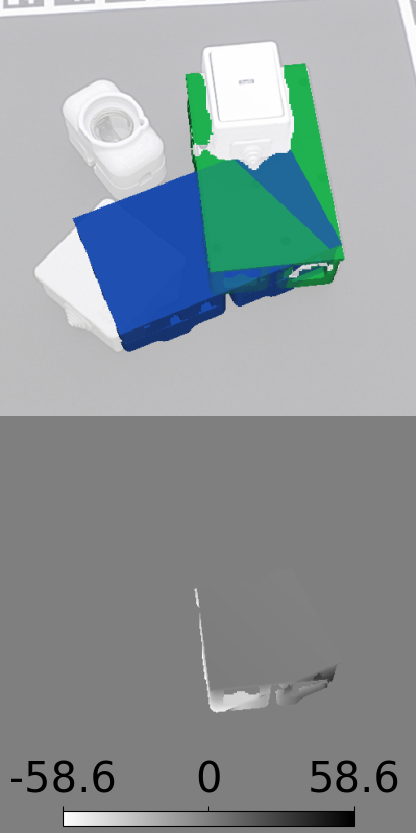} &
            \includegraphics[width=0.117\columnwidth]{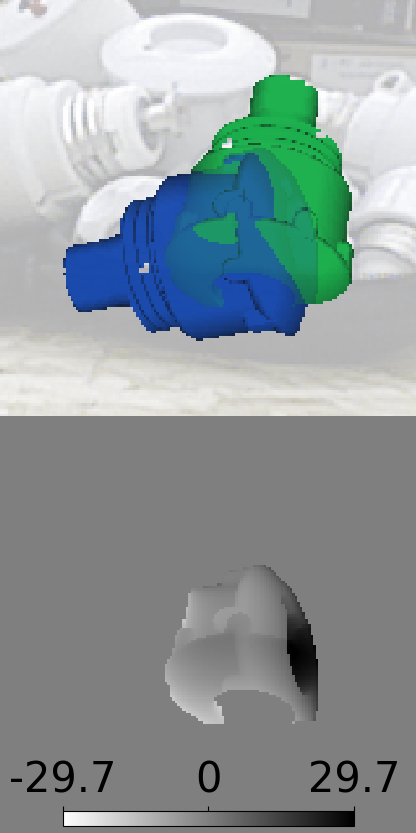} &
            \includegraphics[width=0.117\columnwidth]{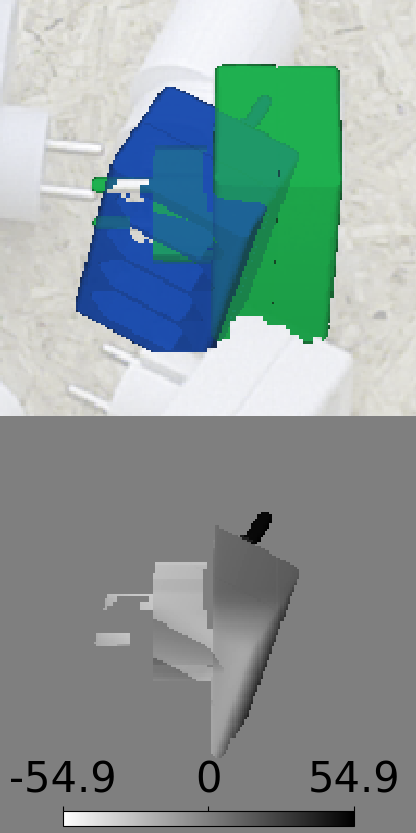} &
            \includegraphics[width=0.117\columnwidth]{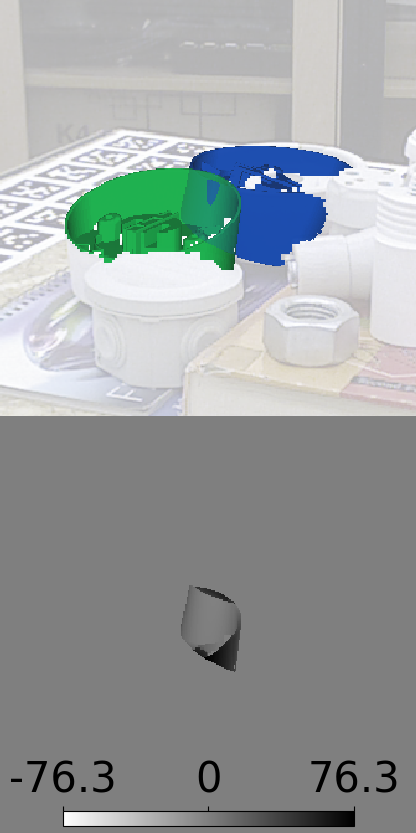} &
            \includegraphics[width=0.117\columnwidth]{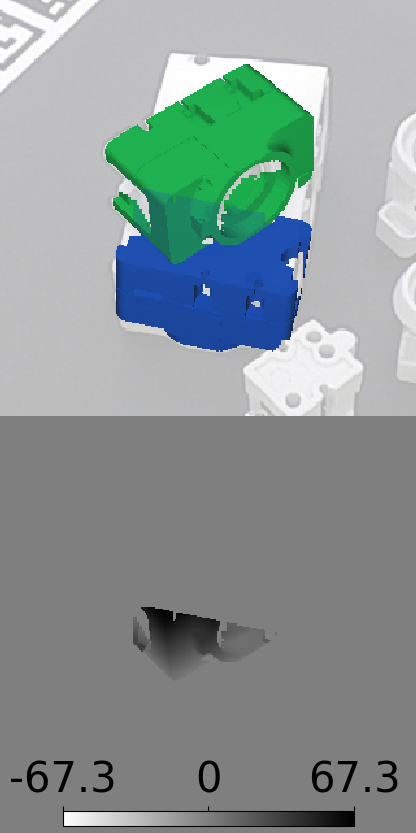} \\
        \end{tabular}
        \caption{Comparison of $e_\mathrm{VSD}$ (bold, $\tau=20\,\si{mm}$) with $e_\mathrm{ADI}/\theta_\mathrm{AD}$ (\si{mm}) on example pose estimates sorted by increasing $e_\mathrm{VSD}$.
            Top: Cropped and brightened test images overlaid with
            renderings of the model at i) the estimated pose $\hat{\mathbf{P}}$ in blue, and ii) the ground-truth pose $\bar{\mathbf{P}}$ in green. Only the part of the model surface that falls into the respective visibility mask is shown. Bottom: Difference maps $S_{\Delta}$.
            Case (b) is analyzed in Fig.~\ref{fig:vsd_components}.
        }
        \label{fig:vsd_examples}
        
        \endgroup
    \end{center}
\end{figure}

\subsubsection{Properties of $e_\mathrm{VSD}$.}

The object pose can be ambiguous, \ie there can be multiple poses that are indistinguishable. This is caused by the existence of multiple fits of the visible part of the object surface to the entire object surface. The visible part is determined by self-occlusion and occlusion by other objects and the multiple surface fits are induced by global or partial object symmetries.

Pose error $e_\mathrm{VSD}$ is calculated only over the visible part of the model surface and thus the indistinguishable poses are treated as equivalent.
This is a desirable property which is not provided by pose-error functions commonly used in the literature~\cite{hodan2016evaluation}, including $e_\mathrm{ADD}$ and $e_\mathrm{ADI}$ discussed below.
As the commonly used pose-error functions, $e_\mathrm{VSD}$ does not consider color information.

Definition (\ref{eq:vsd}) is different from the original definition in~\cite{hodan2016evaluation} where the pixel-wise cost linearly increases to $1$ as $|\hat{S}(p) - \bar{S}(p)|$ increases to~$\tau$. The new definition is easier to interpret
and does not penalize small distance differences that may be caused by imprecisions of the depth sensor or of the ground-truth pose.

\subsubsection{Criterion of Correctness.} An estimated pose $\hat{\mathbf{P}}$ is considered correct w.r.t. the ground-truth pose $\bar{\mathbf{P}}$ if the error $e_\mathrm{VSD} < \theta$. If multiple instances of the target object are visible in the test image, the estimated pose is compared to the ground-truth instance that minimizes the error.
The choice of the misalignment tolerance~$\tau$ and the correctness threshold~$\theta$ depends on the target application. For robotic manipulation, where a robotic arm operates in 3D space, both $\tau$ and $\theta$ need to be low, \eg $\tau=20\,\si{mm}$, $\theta=0.3$, which is the default setting in the evaluation presented in Sec.~\ref{sec:evaluation}. The requirement is different for augmented reality applications.
Here the surface alignment in the $Z$ dimension, \ie the optical axis of the camera, is less important than the alignment
in the $X$ and $Y$ dimension. The tolerance $\tau$ can be therefore relaxed, but $\theta$ needs to stay low.
\pagebreak

\subsubsection{Comparison to Hinterstoisser et al.}

In \cite{hinterstoisser2012accv}, the error is calculated as the average distance from vertices of the model $\mathcal{M}$ in the ground-truth pose $\bar{\mathbf{P}}$ to vertices of $\mathcal{M}$ in the estimated pose $\hat{\mathbf{P}}$. The distance is measured to the position of the same vertex if the object has no indistinguishable views ($e_\mathrm{ADD}$), otherwise to the position of the closest vertex ($e_\mathrm{ADI}$).
The estimated pose~$\hat{\mathbf{P}}$ is considered correct if $e \leq \theta_\mathrm{AD} = 0.1 d$, where $e$ is $e_\mathrm{ADD}$ or $e_\mathrm{ADI}$, and $d$ is the object diameter, \ie the largest distance between any pair of model vertices.

Error $e_\mathrm{ADI}$ can be un-intuitively low because of many-to-one vertex matching established by the search for the closest vertex.
This is shown in Fig.~\ref{fig:vsd_examples}, which compares $e_\mathrm{VSD}$ and $e_\mathrm{ADI}$ on example pose estimates of objects that have indistinguishable views. Overall, (f)-(n) yield low $e_\mathrm{ADI}$ scores and satisfy the correctness criterion of Hinterstoisser et al. These estimates are not considered correct by our criterion.
Estimates (a)-(e) are considered correct and (o)-(p) are considered wrong by both criteria.

\section{Datasets} \label{sec:datasets}

We collected six publicly available datasets, some of which we reduced to remove redundancies\footnote{Identifiers of the selected images are available on the project website.} and re-annotated to ensure a high quality of the ground truth. Additionally, we created two new datasets focusing on varying lighting conditions, since this variation is not present in the existing datasets.
An overview of the datasets is in Fig.~\ref{fig:datasets_overview} and a detailed description follows.

\subsection{Training and Test Data}

The datasets consist of texture-mapped 3D object models and training and test RGB-D images annotated with ground-truth 6D object poses. The 3D object models were created using KinectFusion-like systems for 3D surface reconstruction~\cite{newcombe2011kinectfusion,fastfusion2014}. All images are of approximately VGA resolution.

For training, a method may use the 3D object models and/or the training images.
While 3D models are often available or can be generated at a low cost, capturing and annotating real training images requires a significant effort. The benchmark is therefore focused primarily on the more practical scenario where only the object models, which can be used to render synthetic training images, are available at training time.
All datasets contain already synthesized training images. Methods are allowed to synthesize additional training images, but this option was not utilized for the evaluation in this paper. Only T-LESS and TUD-L include real training images of isolated, \ie non-occluded, objects.

To generate the synthetic training images, objects from the same dataset were rendered from the same range of azimuth/elevation covering the distribution of object poses in the test scenes. The viewpoints were sampled from a sphere, as in~\cite{hinterstoisser2012accv}, with the sphere radius set to the distance of the closest object instance in the test scenes. The objects were rendered with fixed lighting conditions and a black background.

The test images are real images from a structured-light sensor -- Microsoft Kinect v1 or Primesense Carmine 1.09. The test images originate from indoor scenes with varying complexity, ranging from simple scenes with a single isolated object instance to very challenging scenes with multiple instances of several objects and a high amount of clutter and occlusion. Poses of the modeled objects were annotated manually.
While LM, IC-MI and RU-APC provide annotation for instances of only one object per image, the other datasets provide ground-truth for all modeled objects. Details of the datasets are in Tab.~\ref{tab:dataset_params}.

\begin{figure}[!t]
	\begin{center}
		\scriptsize
		\begin{tabularx}{\textwidth}{ l *{8}{R} }
			\toprule
			\multicolumn{1}{l}{{\scriptsize Dataset}} &
			\multicolumn{1}{r}{{\scriptsize Objects}} &
			\multicolumn{2}{r}{{\scriptsize Training images/obj.}} &
			\multicolumn{2}{r}{{\scriptsize Test images}} &
			\multicolumn{2}{r}{{\scriptsize Test targets}} \\
			\cmidrule(l{6pt}r{0pt}){3-4} \cmidrule(l{6pt}r{0pt}){5-6} \cmidrule(l{6pt}r{0pt}){7-8}
			& &
			\multicolumn{1}{r}{{\scriptsize Real}} &
			\multicolumn{1}{r}{{\scriptsize Synt.}} &
			\multicolumn{1}{r}{{\scriptsize Used}} &
			\multicolumn{1}{r}{{\scriptsize All}} &
			\multicolumn{1}{r}{{\scriptsize Used}} &
			\multicolumn{1}{r}{{\scriptsize All}} \\
			\midrule
			LM \cite{hinterstoisser2012accv} & 15 & -- & 1313 & 3000 & 18273 & 3000 & 18273 \\
			LM-O \cite{brachmann2014learning} & 8 & -- & 1313 & 200 & 1214 & 1445 & 8916 \\
			IC-MI \cite{tejani2014latent} & 6 & -- & 1313 & 300 & 2067 & 300 & 2067 \\
			IC-BIN \cite{doumanoglou2016recovering} & 2 & -- & 2377 & 150 & 177 & 200 & 238 \\
			T-LESS \cite{hodan2017tless} & 30 & 1296 & 2562 & 2000 & 10080 & 9819 & 49805 \\
			RU-APC \cite{rennie2016dataset} & 14 & -- & 2562 & 1380 & 5964 & 1380 & 5911\\
			TUD-L - new & 3 & $>$11000 & 1827 & 600 & 23914 & 600 & 23914 \\
			TYO-L - new & 21 & -- & 2562 & -- & 1680 & -- & 1669 \\
			\midrule
			Total & 89 & & & 7450 & 62155 & 16951 & 110793 \\
			\bottomrule
		\end{tabularx}
		\captionof{table}{\label{tab:dataset_params} Parameters of the datasets.
			Note that if a test image shows multiple object models, each model defines a different test target -- see Sec.~\ref{sec:task_formulation}.
		}.
	\end{center}
\end{figure}

\subsection{The Dataset Collection}

\subsubsection{LM/LM-O~\cite{hinterstoisser2012accv,brachmann2014learning}.} LM (a.k.a. Linemod) has been the most commonly used dataset for 6D object pose estimation. It contains 15 texture-less household objects with discriminative color, shape and size. Each object is associated with a test image set showing one annotated object instance with significant
clutter but only mild occlusion. LM-O (a.k.a. Linemod-Occluded) provides ground-truth annotation for all other instances of the modeled objects in one of the test sets. This introduces challenging test cases with various levels of occlusion.

\subsubsection{IC-MI/IC-BIN~\cite{tejani2014latent,doumanoglou2016recovering}.} IC-MI (a.k.a.~Tejani et al.) contains models of two texture-less and four textured household objects. The test images show multiple object instances with clutter and slight occlusion. IC-BIN (a.k.a.~Dou\-ma\-no\-glou et al., scenario 2) includes test images of two objects from IC-MI, which appear in multiple locations with heavy occlusion in a bin-picking scenario. We have removed test images with low-quality ground-truth annotations from both datasets, and refined the annotations for the remaining images in IC-BIN.

\subsubsection{T-LESS~\cite{hodan2017tless}.} It features 30 industry-relevant objects with no significant texture or discriminative color. The objects exhibit symmetries and mutual similarities in shape and/or size, and a few objects are a composition of other objects.
T-LESS includes images from three different sensors and two types of 3D object models. For our evaluation, we only used RGB-D images from the Primesense sensor and the automatically reconstructed 3D object models.

\subsubsection{RU-APC~\cite{rennie2016dataset}.} This dataset (a.k.a. Rutgers APC) includes 14 textured products from the Amazon Picking Challenge 2015~\cite{correll2016lessons}, each associated with test images of a cluttered warehouse shelf. The camera was equipped with LED strips to ensure constant lighting. From the original dataset, we omitted ten objects which are non-rigid or poorly captured by the depth sensor, and included only one from the four images captured from the same viewpoint.

\subsubsection{TUD-L/TYO-L.} Two new datasets with household objects captured under different settings of ambient and directional light. TUD-L (TU Dresden Light) contains training and test image sequences that show three moving objects under eight lighting conditions. The object poses were annotated by manually aligning the 3D object model with the first frame of the sequence and propagating the initial pose through the sequence using ICP. TYO-L (Toyota Light) contains 21 objects, each captured in multiple poses on a table-top setup, with four different table cloths and five different lighting conditions. To obtain the ground truth poses, manually chosen correspondences were utilized to estimate rough poses which were then refined by ICP. The images in both datasets are labeled by categorized lighting conditions.

\section{Evaluated Methods} \label{sec:evaluated_methods}

The evaluated methods cover the major research directions of the 6D object pose estimation field. This section provides a review of the methods, together with a description of the setting of their key parameters. If not stated otherwise, the image-based methods used the synthetic training images.

\subsection{Learning-Based Methods}

\subsubsection{Brachmann-14~\cite{brachmann2014learning}.}
For each pixel of an input image, a regression forest predicts the object identity and
the location in the coordinate frame of the object model,
a so called ``object coordinate''. Simple RGB and depth difference features are used for the prediction.
Each object coordinate prediction defines a 3D-3D correspondence between the image and the 3D object model.
A RANSAC-based optimization schema samples sets of three correspondences to create a pool of pose hypotheses.
The final hypothesis is chosen, and iteratively refined, to maximize the alignment of predicted correspondences, as well as the alignment of observed depth with the object model.
The main parameters of the method were set as follows: maximum feature offset: $20\,\textrm{px}$, features per tree node: 1000, training patches per object: 1.5M, number of trees: 3, size of the hypothesis pool: 210, refined hypotheses: 25. Real training images were used for TUD-L and T-LESS.

\subsubsection{Brachmann-16~\cite{brachmann2016uncertainty}.}

The method of~\cite{brachmann2014learning} is extended in several ways.
Firstly, the random forest is improved using an auto-context algorithm to support pose estimation from RGB-only images.
Secondly, the RANSAC-based optimization hypothesizes not only with regard to the object pose but also with regard to the object identity in cases where it is unknown which objects are visible in the input image.
Both improvements were disabled for the evaluation
since we deal with RGB-D input, and it is known which objects are visible in the image.
Thirdly, the random forest predicts for each pixel a full, three-dimensional distribution over object coordinates capturing uncertainty information.
The distributions are estimated using mean-shift in each forest leaf, and can therefore be heavily multi-modal. 
The final hypothesis is chosen, and iteratively refined, to maximize the likelihood under the predicted distributions.
The 3D object model is not used for fitting the pose.
The parameters were set as: maximum feature offset: $10\,\textrm{px}$, features per tree node: 100, number of trees: 3, number of sampled hypotheses: 256, pixels drawn in each RANSAC iteration: 10K, inlier threshold: $1\,\textrm{cm}$.

\subsubsection{Tejani-14~\cite{tejani2014latent}.}
Linemod~\cite{hinterstoisser2012accv} is adapted into a scale-invariant patch descriptor and integrated into a regression forest with a new template-based split function. This split function is more discriminative than simple pixel tests and accelerated via binary bit-operations. The method is trained on positive samples only, \ie rendered images of the 3D object model. During the inference, the class distributions at the leaf nodes are iteratively updated, providing occlusion-aware segmentation masks.
The object pose is estimated by accumulating pose regression votes from the estimated foreground patches.
The baseline evaluated in this paper implements~\cite{tejani2014latent} but omits the iterative segmentation/refinement step and does not perform ICP. The features and forest parameters were set as in~\cite{tejani2014latent}: number of trees: 10, maximum depth of each tree: 25, number of features in both the color gradient and the surface normal channel: 20, patch size: 1/2 the image, rendered images used to train each forest: 360.

\subsubsection{Kehl-16~\cite{kehl2016deep}.}
Scale-invariant RGB-D patches are extracted from a regular grid attached to the input image, and described by features calculated using a~convolutional auto-encoder.
At training time, a codebook is constructed from descriptors of patches from the training images, with each codebook entry holding information about the 6D pose.
For each patch descriptor from the test image, $k$-nearest neighbors from the codebook are found, and a 6D vote is cast using neighbors whose distance is below a threshold $t$.
After the voting stage, the 6D hypothesis space is filtered to remove spurious votes. Modes are identified by mean-shift and refined by ICP. The final hypothesis is verified in color, depth and surface normals to suppress false positives. The main parameters of the method with the used values: patch size: $32\times32\,\textrm{px}$, patch sampling step:
$6\,\textrm{px}$, $k$-nearest neighbors:~3, threshold $t$: $2$, number of extracted modes from the pose space: 8. Real training images were used for T-LESS.

\subsection{Template Matching Methods}

\subsubsection{Hodaň-15~\cite{hodan2015detection}.}

A template matching method that applies an efficient cascade-style evaluation to each sliding window location. A simple objectness filter is applied first, rapidly rejecting most locations. For each remaining location, a~set of candidate templates is identified by a voting procedure based on hashing, which makes the computational complexity largely unaffected by the total number of stored templates. The candidate templates are then verified as in Linemod~\cite{hinterstoisser2012accv} by matching feature points in different modalities (surface normals, image gradients, depth, color). Finally, object poses associated with the detected templates are refined by particle swarm optimization (PSO). The templates were generated by applying the full circle of in-plane rotations with $10^\circ$ step to a portion of the synthetic training images, resulting in 11--23K templates per object. Other parameters were set as described in~\cite{hodan2015detection}.  We present also results  without the last refinement step (Hodaň-15-nr).

\subsection{Methods Based on Point-Pair Features}

\subsubsection{Drost-10~\cite{drost2010model}.}
A method based on matching oriented point pairs between the point cloud of the test scene and the object model, and grouping the matches using a local voting scheme. 
At training time, point pairs from the model are sampled and stored in a hash table.
At test time, reference points are fixed in the scene, and a low-dimensional parameter space for the voting scheme is created by restricting to those poses that align the reference point with the model. Point pairs between the reference point and other scene points are created, similar model point pairs searched for using the hash table, and a vote is cast for each matching point pair.
Peaks in the accumulator space are extracted and used as pose candidates, which are refined by coarse-to-fine ICP and re-scored by the relative amount of visible model surface.
Note that color information is not used. It was evaluated using function \texttt{find\_surface\_model} from HALCON 13.0.2~\cite{mvtechalcon}. The sampling distances for model and scene were set to 3\% of the object diameter, 10\% of points were used as the reference points, and the normals were computed using the \texttt{mls} method. Points further than 2\,\si{m} were discarded.

\subsubsection{Drost-10-edge.}
An extension of~\cite{drost2010model} which additionally detects 3D edges from the scene and favors poses in which the model contours are aligned with the edges. A multi-modal refinement minimizes the surface distances and the distances of reprojected model contours to the detected edges. The evaluation was performed using the same software and parameters as Drost-10, but with activated parameter \texttt{train\_3d\_edges} during the model creation.

\subsubsection{Vidal-18~\cite{vidal2018sixd}.}
The point cloud is first sub-sampled by clustering points based~on the surface normal orientation.
Inspired by improvements of~\cite{hinterstoisser2016going}, the matching strategy of~\cite{drost2010model} was improved by mitigating the effect of the feature discretization step. Additionally, an improved non-maximum suppression of the pose candidates from different reference points removes spurious matches.
The most voted 500 pose candidates are sorted by a surface fitting score and the 200 best candidates are refined by projective ICP.
For the final 10 candidates, the consistency of the object surface and silhouette with the scene is evaluated.
The sampling distance for model, scene and features was set to 5\% of the object diameter, and 20\%~of the scene points were used as the reference points.

\subsection{Methods Based on 3D Local Features}

\subsubsection{Buch-16~\cite{buch2016local}.}

A RANSAC-based method that iteratively samples three feature correspondences between the object model and the scene. The correspondences are obtained by matching 3D local shape descriptors and are used to generate a 6D pose candidate, whose quality is measured by the consensus set size. The final pose is refined by ICP.
The method achieved the state-of-the-art results on earlier object recognition datasets captured by LIDAR, but suffers from a cubic complexity in the number of correspondences. The number of RANSAC iterations was set to 10000, allowing only for a limited search in cluttered scenes. The method was evaluated with several descriptors:
153d SI~\cite{johnson1999using},
352d SHOT~\cite{salti2014shot},
30d ECSAD~\cite{jorgensen2015geometric},
and 1536d PPFH~\cite{buch2018local}.
None of the descriptors utilize color.

\subsubsection{Buch-17~\cite{buch2017rotational}.}
This method is based on the observation that a correspondence between two oriented points on the object surface is constrained to cast votes in a~1-DoF rotational subgroup of the full group of poses, SE(3). The time complexity of the method is thus linear in the number of correspondences. Kernel density estimation is used to efficiently combine the votes and generate a 6D pose estimate. As Buch-16, the method relies on 3D local shape descriptors and refines the final pose estimate by ICP.
The parameters were set as in the paper: 60~angle tessellations were used for casting rotational votes, and  the translation/rotation bandwidths were set to 10\,\si{mm}/22.5$^\circ$.

\section{Evaluation} \label{sec:evaluation}

The methods reviewed in Sec.~\ref{sec:evaluated_methods} were evaluated by their original authors on the datasets described in Sec.~\ref{sec:datasets}, using the evaluation methodology from Sec.~\ref{sec:methodology}.

\subsection{Experimental Setup}

\subsubsection{Fixed Parameters.}
The parameters of each method were fixed for all objects and datasets.
The distribution of object poses in the test scenes was the only dataset-specific information used by the methods. The distribution determined the range of viewpoints from which the object models were rendered to obtain synthetic training images.

\subsubsection{Pose Error.}
The error of a 6D object pose estimate is measured with the pose-error function $e_\mathrm{VSD}$ defined in Sec.~\ref{sec:error}. The visibility masks were calculated as in~\cite{hodan2016evaluation}, with the occlusion tolerance $\delta$ set to $15\,\si{mm}$. Only the ground truth poses in which the object is visible from at least 10\% were considered in the evaluation.

\subsubsection{Performance Score.}
The performance is measured by the recall score, \ie the fraction of test targets for which a correct object pose was estimated. Recall scores per dataset and per object are reported. The overall performance is given by the average of per-dataset recall scores. We thus treat each dataset as a separate challenge and avoid the overall score being dominated by larger datasets.

\subsubsection{Subsets Used for the Evaluation.}
We reduced the number of test images to remove redundancies and to encourage participation of new, in particular slow, methods.
From the total of 62K test images, we sub-sampled 7K, reducing the number of test targets from 110K to 17K (Tab.~\ref{tab:dataset_params}).
Full datasets with identifiers of the selected test images are on the project website.
TYO-L was not used for the evaluation presented in this paper, but it is a part of the online evaluation.

\begin{figure}[!t]
	\begin{center}
		\tiny
		\begin{tabularx}{\textwidth}{ r l *{7}{R} R | R }
			\toprule
			\multicolumn{1}{r}{{\tiny \#}} &
			\multicolumn{1}{l}{{\tiny Method}} &
			\multicolumn{1}{c}{{\tiny LM}} &
			\multicolumn{1}{c}{{\tiny LM-O}} &
			\multicolumn{1}{c}{{\tiny IC-MI}} &
			\multicolumn{1}{c}{{\tiny IC-BIN}} &
			\multicolumn{1}{c}{{\tiny T-LESS}} &
			\multicolumn{1}{c}{{\tiny RU-APC}} &
			\multicolumn{1}{c}{{\tiny TUD-L}} &
			\multicolumn{1}{c}{{\tiny Average}} &
			\multicolumn{1}{c}{{\tiny Time (s)}} \\
			
			\midrule
			
			1. & Vidal-18 & \cellcolor{ccol!26}87.83 & \cellcolor{ccol!18}59.31 & \cellcolor{ccol!29}95.33 & \cellcolor{ccol!29}96.50 & \cellcolor{ccol!20}66.51 & \cellcolor{ccol!11}36.52 & \cellcolor{ccol!24}80.17 & \cellcolor{ccol!22}74.60 & 4.7 \\
			2. & Drost-10-edge & \cellcolor{ccol!24}79.13 & \cellcolor{ccol!16}54.95 & \cellcolor{ccol!28}94.00 & \cellcolor{ccol!28}92.00 & \cellcolor{ccol!20}67.50 & \cellcolor{ccol!8}27.17 & \cellcolor{ccol!26}87.33 & \cellcolor{ccol!22}71.73 & 21.5 \\
			3. & Drost-10 & \cellcolor{ccol!25}82.00 & \cellcolor{ccol!17}55.36 & \cellcolor{ccol!28}94.33 & \cellcolor{ccol!26}87.00 & \cellcolor{ccol!17}56.81 & \cellcolor{ccol!7}22.25 & \cellcolor{ccol!24}78.67 & \cellcolor{ccol!20}68.06 & 2.3 \\
			4. & Hodan-15 & \cellcolor{ccol!26}87.10 & \cellcolor{ccol!15}51.42 & \cellcolor{ccol!29}95.33 & \cellcolor{ccol!27}90.50 & \cellcolor{ccol!19}63.18 & \cellcolor{ccol!11}37.61 & \cellcolor{ccol!14}45.50 & \cellcolor{ccol!20}67.23 & 13.5 \\
			5. & Brachmann-16 & \cellcolor{ccol!23}75.33 & \cellcolor{ccol!16}52.04 & \cellcolor{ccol!22}73.33 & \cellcolor{ccol!17}56.50 & \cellcolor{ccol!5}17.84 & \cellcolor{ccol!7}24.35 & \cellcolor{ccol!27}88.67 & \cellcolor{ccol!17}55.44 & 4.4 \\
			6. & Hodan-15-nopso & \cellcolor{ccol!21}69.83 & \cellcolor{ccol!10}34.39 & \cellcolor{ccol!25}84.67 & \cellcolor{ccol!23}76.00 & \cellcolor{ccol!19}62.70 & \cellcolor{ccol!10}32.39 & \cellcolor{ccol!8}27.83 & \cellcolor{ccol!17}55.40 & 12.3 \\
			7. & Buch-17-ppfh & \cellcolor{ccol!17}56.60 & \cellcolor{ccol!11}36.96 & \cellcolor{ccol!29}95.00 & \cellcolor{ccol!23}75.00 & \cellcolor{ccol!8}25.10 & \cellcolor{ccol!6}20.80 & \cellcolor{ccol!21}68.67 & \cellcolor{ccol!16}54.02 & 14.2 \\
			8. & Kehl-16 & \cellcolor{ccol!17}58.20 & \cellcolor{ccol!10}33.91 & \cellcolor{ccol!20}65.00 & \cellcolor{ccol!13}44.00 & \cellcolor{ccol!7}24.60 & \cellcolor{ccol!8}25.58 & \cellcolor{ccol!2}7.50 & \cellcolor{ccol!11}36.97 & 1.8 \\
			9. & Buch-17-si & \cellcolor{ccol!10}33.33 & \cellcolor{ccol!6}20.35 & \cellcolor{ccol!20}67.33 & \cellcolor{ccol!18}59.00 & \cellcolor{ccol!4}13.34 & \cellcolor{ccol!7}23.12 & \cellcolor{ccol!12}41.17 & \cellcolor{ccol!11}36.81 & 15.9 \\
			10. & Brachmann-14 & \cellcolor{ccol!20}67.60 & \cellcolor{ccol!12}41.52 & \cellcolor{ccol!24}78.67 & \cellcolor{ccol!7}24.00 & \cellcolor{ccol!0}0.25 & \cellcolor{ccol!9}30.22 & \cellcolor{ccol!0}0.00 & \cellcolor{ccol!10}34.61 & 1.4 \\
			11. & Buch-17-ecsad & \cellcolor{ccol!4}13.27 & \cellcolor{ccol!3}9.62 & \cellcolor{ccol!12}40.67 & \cellcolor{ccol!18}59.00 & \cellcolor{ccol!2}7.16 & \cellcolor{ccol!2}6.59 & \cellcolor{ccol!7}24.00 & \cellcolor{ccol!7}22.90 & 5.9 \\
			12. & Buch-17-shot & \cellcolor{ccol!2}5.97 & \cellcolor{ccol!0}1.45 & \cellcolor{ccol!13}43.00 & \cellcolor{ccol!12}38.50 & \cellcolor{ccol!1}3.83 & \cellcolor{ccol!0}0.07 & \cellcolor{ccol!5}16.67 & \cellcolor{ccol!5}15.64 & 6.7 \\
			13. & Tejani-14 & \cellcolor{ccol!4}12.10 & \cellcolor{ccol!1}4.50 & \cellcolor{ccol!11}36.33 & \cellcolor{ccol!3}10.00 & \cellcolor{ccol!0}0.13 & \cellcolor{ccol!0}1.52 & \cellcolor{ccol!0}0.00 & \cellcolor{ccol!3}9.23 & 1.4 \\
			14. & Buch-16-ppfh & \cellcolor{ccol!2}8.13 & \cellcolor{ccol!1}2.28 & \cellcolor{ccol!6}20.00 & \cellcolor{ccol!1}2.50 & \cellcolor{ccol!2}7.81 & \cellcolor{ccol!3}8.99 & \cellcolor{ccol!0}0.67 & \cellcolor{ccol!2}7.20 & 47.1 \\
			15. & Buch-16-ecsad & \cellcolor{ccol!1}3.70 & \cellcolor{ccol!0}0.97 & \cellcolor{ccol!1}3.67 & \cellcolor{ccol!1}4.00 & \cellcolor{ccol!0}1.24 & \cellcolor{ccol!1}2.90 & \cellcolor{ccol!0}0.17 & \cellcolor{ccol!1}2.38 & 39.1 \\
			
			\bottomrule
		\end{tabularx}
		
		\captionof{table}{\label{tab:eval_per_dataset} Recall scores (\%) for $\tau=20\,\si{mm}$ and $\theta=0.3$. The recall score is the percentage of test targets
			for which a~correct object pose was estimated. The methods are sorted by their average recall score calculated as the average of the per-dataset recall scores. The right-most column shows the average running time per test target.
		}
		
		\vspace{10ex}
		
		\tiny
		\begin{tabularx}{\textwidth}{ r l *{26}{R}}
			\toprule
			\# & {\tiny Method} & \multicolumn{15}{c}{{\tiny LM}} & \multicolumn{8}{c}{{\tiny LM-O}} & \multicolumn{3}{c}{{\tiny TUD-L}} \\
			\cmidrule(l{2pt}r{2pt}){3-17} \cmidrule(l{2pt}r{2pt}){18-25} \cmidrule(l{2pt}r{2pt}){26-28}
			& & 1 & 2 & 3 & 4 & 5 & 6 & 7 & 8 & 9 & 10 & 11 & 12 & 13 & 14 & 15 & 1 & 5 & 6 & 8 & 9 & 10 & 11 & 12 & 1 & 2 & 3 \\
			\midrule
			
			1. & Vidal-18 & \cellcolor{ccol!27}89 & \cellcolor{ccol!29}96 & \cellcolor{ccol!27}91 & \cellcolor{ccol!28}94 & \cellcolor{ccol!27}92 & \cellcolor{ccol!29}96 & \cellcolor{ccol!27}89 & \cellcolor{ccol!27}89 & \cellcolor{ccol!26}87 & \cellcolor{ccol!29}97 & \cellcolor{ccol!18}59 & \cellcolor{ccol!21}69 & \cellcolor{ccol!28}93 & \cellcolor{ccol!28}92 & \cellcolor{ccol!27}90 & \cellcolor{ccol!20}66 & \cellcolor{ccol!24}81 & \cellcolor{ccol!14}46 & \cellcolor{ccol!19}65 & \cellcolor{ccol!22}73 & \cellcolor{ccol!13}43 & \cellcolor{ccol!8}26 & \cellcolor{ccol!19}64 & \cellcolor{ccol!24}79 & \cellcolor{ccol!26}88 & \cellcolor{ccol!22}74 \\
			2. & Drost-10-edge & \cellcolor{ccol!23}77 & \cellcolor{ccol!29}97 & \cellcolor{ccol!28}94 & \cellcolor{ccol!12}40 & \cellcolor{ccol!29}98 & \cellcolor{ccol!28}94 & \cellcolor{ccol!25}83 & \cellcolor{ccol!29}96 & \cellcolor{ccol!14}45 & \cellcolor{ccol!28}94 & \cellcolor{ccol!20}68 & \cellcolor{ccol!20}66 & \cellcolor{ccol!21}72 & \cellcolor{ccol!26}88 & \cellcolor{ccol!24}79 & \cellcolor{ccol!14}47 & \cellcolor{ccol!25}82 & \cellcolor{ccol!14}46 & \cellcolor{ccol!22}75 & \cellcolor{ccol!13}42 & \cellcolor{ccol!13}44 & \cellcolor{ccol!11}36 & \cellcolor{ccol!17}57 & \cellcolor{ccol!26}85 & \cellcolor{ccol!26}88 & \cellcolor{ccol!27}90 \\
			3. & Drost-10 & \cellcolor{ccol!26}86 & \cellcolor{ccol!25}83 & \cellcolor{ccol!27}89 & \cellcolor{ccol!25}84 & \cellcolor{ccol!28}93 & \cellcolor{ccol!26}87 & \cellcolor{ccol!26}86 & \cellcolor{ccol!28}92 & \cellcolor{ccol!20}66 & \cellcolor{ccol!29}96 & \cellcolor{ccol!16}53 & \cellcolor{ccol!20}67 & \cellcolor{ccol!24}79 & \cellcolor{ccol!27}91 & \cellcolor{ccol!24}80 & \cellcolor{ccol!19}62 & \cellcolor{ccol!22}75 & \cellcolor{ccol!12}39 & \cellcolor{ccol!21}70 & \cellcolor{ccol!17}57 & \cellcolor{ccol!14}46 & \cellcolor{ccol!8}26 & \cellcolor{ccol!17}57 & \cellcolor{ccol!22}73 & \cellcolor{ccol!27}90 & \cellcolor{ccol!22}74 \\
			4. & Hodan-15 & \cellcolor{ccol!27}91 & \cellcolor{ccol!29}97 & \cellcolor{ccol!24}79 & \cellcolor{ccol!29}97 & \cellcolor{ccol!27}91 & \cellcolor{ccol!29}97 & \cellcolor{ccol!22}73 & \cellcolor{ccol!21}69 & \cellcolor{ccol!27}90 & \cellcolor{ccol!29}97 & \cellcolor{ccol!24}81 & \cellcolor{ccol!24}79 & \cellcolor{ccol!30}99 & \cellcolor{ccol!22}74 & \cellcolor{ccol!28}95 & \cellcolor{ccol!16}54 & \cellcolor{ccol!20}66 & \cellcolor{ccol!12}40 & \cellcolor{ccol!8}26 & \cellcolor{ccol!22}73 & \cellcolor{ccol!11}37 & \cellcolor{ccol!13}44 & \cellcolor{ccol!20}68 & \cellcolor{ccol!8}27 & \cellcolor{ccol!19}63 & \cellcolor{ccol!14}48 \\
			5. & Brachmann-16 & \cellcolor{ccol!27}92 & \cellcolor{ccol!28}93 & \cellcolor{ccol!23}76 & \cellcolor{ccol!25}84 & \cellcolor{ccol!26}86 & \cellcolor{ccol!27}90 & \cellcolor{ccol!13}44 & \cellcolor{ccol!22}72 & \cellcolor{ccol!25}85 & \cellcolor{ccol!24}79 & \cellcolor{ccol!14}46 & \cellcolor{ccol!20}67 & \cellcolor{ccol!28}94 & \cellcolor{ccol!18}60 & \cellcolor{ccol!20}66 & \cellcolor{ccol!19}64 & \cellcolor{ccol!20}65 & \cellcolor{ccol!13}44 & \cellcolor{ccol!20}68 & \cellcolor{ccol!21}71 & \cellcolor{ccol!1}3 & \cellcolor{ccol!10}32 & \cellcolor{ccol!18}61 & \cellcolor{ccol!24}81 & \cellcolor{ccol!28}95 & \cellcolor{ccol!27}91 \\
			6. & Hodan-15-nr & \cellcolor{ccol!27}91 & \cellcolor{ccol!17}57 & \cellcolor{ccol!12}40 & \cellcolor{ccol!27}89 & \cellcolor{ccol!20}66 & \cellcolor{ccol!26}87 & \cellcolor{ccol!18}59 & \cellcolor{ccol!15}49 & \cellcolor{ccol!27}92 & \cellcolor{ccol!27}90 & \cellcolor{ccol!20}65 & \cellcolor{ccol!19}63 & \cellcolor{ccol!21}71 & \cellcolor{ccol!16}54 & \cellcolor{ccol!24}79 & \cellcolor{ccol!14}47 & \cellcolor{ccol!11}35 & \cellcolor{ccol!7}24 & \cellcolor{ccol!3}12 & \cellcolor{ccol!19}63 & \cellcolor{ccol!3}9 & \cellcolor{ccol!10}32 & \cellcolor{ccol!16}53 & \cellcolor{ccol!3}12 & \cellcolor{ccol!16}52 & \cellcolor{ccol!6}20 \\
			7. & Buch-17-ppfh & \cellcolor{ccol!23}77 & \cellcolor{ccol!20}65 & \cellcolor{ccol!0}0 & \cellcolor{ccol!28}94 & \cellcolor{ccol!25}84 & \cellcolor{ccol!18}60 & \cellcolor{ccol!7}24 & \cellcolor{ccol!18}59 & \cellcolor{ccol!22}75 & \cellcolor{ccol!20}67 & \cellcolor{ccol!7}24 & \cellcolor{ccol!12}39 & \cellcolor{ccol!22}75 & \cellcolor{ccol!14}47 & \cellcolor{ccol!19}62 & \cellcolor{ccol!18}59 & \cellcolor{ccol!19}63 & \cellcolor{ccol!5}18 & \cellcolor{ccol!10}35 & \cellcolor{ccol!18}60 & \cellcolor{ccol!5}17 & \cellcolor{ccol!2}5 & \cellcolor{ccol!9}30 & \cellcolor{ccol!16}55 & \cellcolor{ccol!27}89 & \cellcolor{ccol!19}63 \\
			8. & Kehl-16 & \cellcolor{ccol!18}60 & \cellcolor{ccol!16}52 & \cellcolor{ccol!24}81 & \cellcolor{ccol!7}25 & \cellcolor{ccol!24}79 & \cellcolor{ccol!20}68 & \cellcolor{ccol!5}17 & \cellcolor{ccol!20}68 & \cellcolor{ccol!12}42 & \cellcolor{ccol!27}91 & \cellcolor{ccol!13}45 & \cellcolor{ccol!12}42 & \cellcolor{ccol!23}78 & \cellcolor{ccol!25}83 & \cellcolor{ccol!14}46 & \cellcolor{ccol!12}39 & \cellcolor{ccol!14}47 & \cellcolor{ccol!7}24 & \cellcolor{ccol!9}30 & \cellcolor{ccol!15}48 & \cellcolor{ccol!4}14 & \cellcolor{ccol!4}13 & \cellcolor{ccol!15}49 & \cellcolor{ccol!0}0 & \cellcolor{ccol!7}23 & \cellcolor{ccol!0}0 \\
			9. & Buch-17-si & \cellcolor{ccol!12}40 & \cellcolor{ccol!13}43 & \cellcolor{ccol!0}1 & \cellcolor{ccol!19}63 & \cellcolor{ccol!24}81 & \cellcolor{ccol!14}47 & \cellcolor{ccol!4}12 & \cellcolor{ccol!2}8 & \cellcolor{ccol!11}36 & \cellcolor{ccol!13}43 & \cellcolor{ccol!5}18 & \cellcolor{ccol!1}3 & \cellcolor{ccol!14}46 & \cellcolor{ccol!6}19 & \cellcolor{ccol!13}43 & \cellcolor{ccol!16}54 & \cellcolor{ccol!19}63 & \cellcolor{ccol!3}11 & \cellcolor{ccol!0}2 & \cellcolor{ccol!5}16 & \cellcolor{ccol!3}9 & \cellcolor{ccol!0}1 & \cellcolor{ccol!1}3 & \cellcolor{ccol!1}2 & \cellcolor{ccol!22}74 & \cellcolor{ccol!14}48 \\
			10. & Brachmann-14 & \cellcolor{ccol!22}74 & \cellcolor{ccol!21}70 & \cellcolor{ccol!23}77 & \cellcolor{ccol!23}75 & \cellcolor{ccol!26}88 & \cellcolor{ccol!20}66 & \cellcolor{ccol!3}11 & \cellcolor{ccol!24}81 & \cellcolor{ccol!21}69 & \cellcolor{ccol!20}66 & \cellcolor{ccol!15}50 & \cellcolor{ccol!22}75 & \cellcolor{ccol!27}92 & \cellcolor{ccol!23}75 & \cellcolor{ccol!15}49 & \cellcolor{ccol!15}50 & \cellcolor{ccol!14}48 & \cellcolor{ccol!8}27 & \cellcolor{ccol!13}44 & \cellcolor{ccol!18}60 & \cellcolor{ccol!2}6 & \cellcolor{ccol!9}30 & \cellcolor{ccol!19}62 & \cellcolor{ccol!0}0 & \cellcolor{ccol!0}0 & \cellcolor{ccol!0}0 \\
			11. & Buch-17-ecsad & \cellcolor{ccol!9}31 & \cellcolor{ccol!1}2 & \cellcolor{ccol!1}2 & \cellcolor{ccol!6}19 & \cellcolor{ccol!20}66 & \cellcolor{ccol!1}3 & \cellcolor{ccol!1}3 & \cellcolor{ccol!0}0 & \cellcolor{ccol!3}9 & \cellcolor{ccol!15}49 & \cellcolor{ccol!0}1 & \cellcolor{ccol!0}0 & \cellcolor{ccol!1}3 & \cellcolor{ccol!2}7 & \cellcolor{ccol!2}6 & \cellcolor{ccol!9}29 & \cellcolor{ccol!9}29 & \cellcolor{ccol!0}0 & \cellcolor{ccol!0}0 & \cellcolor{ccol!2}7 & \cellcolor{ccol!3}8 & \cellcolor{ccol!0}1 & \cellcolor{ccol!0}0 & \cellcolor{ccol!0}1 & \cellcolor{ccol!19}62 & \cellcolor{ccol!3}10 \\
			12. & Buch-17-shot & \cellcolor{ccol!1}3 & \cellcolor{ccol!1}4 & \cellcolor{ccol!3}11 & \cellcolor{ccol!3}9 & \cellcolor{ccol!3}9 & \cellcolor{ccol!1}4 & \cellcolor{ccol!0}1 & \cellcolor{ccol!1}3 & \cellcolor{ccol!0}2 & \cellcolor{ccol!3}10 & \cellcolor{ccol!0}1 & \cellcolor{ccol!0}0 & \cellcolor{ccol!3}10 & \cellcolor{ccol!4}12 & \cellcolor{ccol!4}14 & \cellcolor{ccol!1}2 & \cellcolor{ccol!2}7 & \cellcolor{ccol!0}0 & \cellcolor{ccol!0}0 & \cellcolor{ccol!0}1 & \cellcolor{ccol!0}1 & \cellcolor{ccol!0}1 & \cellcolor{ccol!0}0 & \cellcolor{ccol!0}1 & \cellcolor{ccol!10}33 & \cellcolor{ccol!5}17 \\
			13. & Tejani-14 & \cellcolor{ccol!11}36 & \cellcolor{ccol!0}0 & \cellcolor{ccol!11}36 & \cellcolor{ccol!0}0 & \cellcolor{ccol!0}1 & \cellcolor{ccol!0}0 & \cellcolor{ccol!0}1 & \cellcolor{ccol!3}11 & \cellcolor{ccol!0}1 & \cellcolor{ccol!21}70 & \cellcolor{ccol!8}27 & \cellcolor{ccol!0}0 & \cellcolor{ccol!0}0 & \cellcolor{ccol!0}0 & \cellcolor{ccol!0}0 & \cellcolor{ccol!8}26 & \cellcolor{ccol!1}2 & \cellcolor{ccol!0}0 & \cellcolor{ccol!0}1 & \cellcolor{ccol!0}0 & \cellcolor{ccol!0}0 & \cellcolor{ccol!3}10 & \cellcolor{ccol!0}0 & \cellcolor{ccol!0}0 & \cellcolor{ccol!0}0 & \cellcolor{ccol!0}0 \\
			14. & Buch-16-ppfh & \cellcolor{ccol!3}11 & \cellcolor{ccol!0}0 & \cellcolor{ccol!0}1 & \cellcolor{ccol!6}22 & \cellcolor{ccol!1}3 & \cellcolor{ccol!2}7 & \cellcolor{ccol!0}2 & \cellcolor{ccol!2}7 & \cellcolor{ccol!5}18 & \cellcolor{ccol!4}12 & \cellcolor{ccol!1}4 & \cellcolor{ccol!1}3 & \cellcolor{ccol!3}9 & \cellcolor{ccol!4}12 & \cellcolor{ccol!4}14 & \cellcolor{ccol!1}4 & \cellcolor{ccol!0}0 & \cellcolor{ccol!0}0 & \cellcolor{ccol!1}2 & \cellcolor{ccol!3}11 & \cellcolor{ccol!0}1 & \cellcolor{ccol!0}1 & \cellcolor{ccol!0}1 & \cellcolor{ccol!1}2 & \cellcolor{ccol!0}0 & \cellcolor{ccol!0}0 \\
			15. & Buch-16-ecsad & \cellcolor{ccol!1}2 & \cellcolor{ccol!0}0 & \cellcolor{ccol!0}0 & \cellcolor{ccol!3}9 & \cellcolor{ccol!1}5 & \cellcolor{ccol!0}0 & \cellcolor{ccol!0}0 & \cellcolor{ccol!1}4 & \cellcolor{ccol!2}5 & \cellcolor{ccol!2}8 & \cellcolor{ccol!0}0 & \cellcolor{ccol!0}0 & \cellcolor{ccol!5}17 & \cellcolor{ccol!1}3 & \cellcolor{ccol!2}5 & \cellcolor{ccol!0}1 & \cellcolor{ccol!1}3 & \cellcolor{ccol!0}0 & \cellcolor{ccol!1}2 & \cellcolor{ccol!1}2 & \cellcolor{ccol!0}0 & \cellcolor{ccol!0}0 & \cellcolor{ccol!0}0 & \cellcolor{ccol!0}0 & \cellcolor{ccol!0}1 & \cellcolor{ccol!0}0 \\

			\toprule
			
			& {\tiny } & \multicolumn{6}{c}{{\tiny IC-MI}} & \multicolumn{2}{c}{{\tiny -BIN}} & \multicolumn{18}{c}{{\tiny T-LESS}} \\
			\cmidrule(l{2pt}r{2pt}){3-8} \cmidrule(l{2pt}r{2pt}){9-10} \cmidrule(l{2pt}r{2pt}){11-28}
			& & 1 & 2 & 3 & 4 & 5 & 6 & 2 & 4 & 1 & 2 & 3 & 4 & 5 & 6 & 7 & 8 & 9 & 10 & 11 & 12 & 13 & 14 & 15 & 16 & 17 & 18 \\
			\midrule
			
			1. & Vidal-18 & \cellcolor{ccol!24}80 & \cellcolor{ccol!30}100 & \cellcolor{ccol!30}100 & \cellcolor{ccol!29}98 & \cellcolor{ccol!30}100 & \cellcolor{ccol!28}94 & \cellcolor{ccol!30}100 & \cellcolor{ccol!28}93 & \cellcolor{ccol!13}43 & \cellcolor{ccol!14}46 & \cellcolor{ccol!20}68 & \cellcolor{ccol!20}65 & \cellcolor{ccol!21}69 & \cellcolor{ccol!21}71 & \cellcolor{ccol!23}76 & \cellcolor{ccol!23}76 & \cellcolor{ccol!28}92 & \cellcolor{ccol!21}69 & \cellcolor{ccol!21}68 & \cellcolor{ccol!25}84 & \cellcolor{ccol!17}55 & \cellcolor{ccol!14}47 & \cellcolor{ccol!16}54 & \cellcolor{ccol!25}85 & \cellcolor{ccol!25}82 & \cellcolor{ccol!24}79 \\
			2. & Drost-10-edge & \cellcolor{ccol!23}78 & \cellcolor{ccol!30}100 & \cellcolor{ccol!30}100 & \cellcolor{ccol!30}100 & \cellcolor{ccol!27}90 & \cellcolor{ccol!29}96 & \cellcolor{ccol!30}100 & \cellcolor{ccol!25}84 & \cellcolor{ccol!16}53 & \cellcolor{ccol!13}44 & \cellcolor{ccol!18}61 & \cellcolor{ccol!20}67 & \cellcolor{ccol!21}71 & \cellcolor{ccol!22}73 & \cellcolor{ccol!23}75 & \cellcolor{ccol!27}89 & \cellcolor{ccol!28}92 & \cellcolor{ccol!22}72 & \cellcolor{ccol!19}64 & \cellcolor{ccol!24}81 & \cellcolor{ccol!16}53 & \cellcolor{ccol!14}46 & \cellcolor{ccol!16}55 & \cellcolor{ccol!25}85 & \cellcolor{ccol!26}88 & \cellcolor{ccol!23}78 \\
			3. & Drost-10 & \cellcolor{ccol!23}76 & \cellcolor{ccol!30}100 & \cellcolor{ccol!29}98 & \cellcolor{ccol!30}100 & \cellcolor{ccol!29}96 & \cellcolor{ccol!29}96 & \cellcolor{ccol!30}100 & \cellcolor{ccol!22}74 & \cellcolor{ccol!10}34 & \cellcolor{ccol!14}46 & \cellcolor{ccol!19}63 & \cellcolor{ccol!19}63 & \cellcolor{ccol!20}68 & \cellcolor{ccol!19}64 & \cellcolor{ccol!16}54 & \cellcolor{ccol!14}48 & \cellcolor{ccol!18}59 & \cellcolor{ccol!16}54 & \cellcolor{ccol!15}51 & \cellcolor{ccol!21}69 & \cellcolor{ccol!13}43 & \cellcolor{ccol!13}45 & \cellcolor{ccol!16}53 & \cellcolor{ccol!24}80 & \cellcolor{ccol!24}79 & \cellcolor{ccol!20}68 \\
			4. & Hodan-15 & \cellcolor{ccol!30}100 & \cellcolor{ccol!30}100 & \cellcolor{ccol!30}100 & \cellcolor{ccol!22}74 & \cellcolor{ccol!29}98 & \cellcolor{ccol!30}100 & \cellcolor{ccol!30}100 & \cellcolor{ccol!24}81 & \cellcolor{ccol!20}66 & \cellcolor{ccol!20}67 & \cellcolor{ccol!22}72 & \cellcolor{ccol!22}72 & \cellcolor{ccol!18}61 & \cellcolor{ccol!18}60 & \cellcolor{ccol!16}52 & \cellcolor{ccol!18}61 & \cellcolor{ccol!26}86 & \cellcolor{ccol!22}72 & \cellcolor{ccol!17}56 & \cellcolor{ccol!17}55 & \cellcolor{ccol!16}54 & \cellcolor{ccol!6}21 & \cellcolor{ccol!18}59 & \cellcolor{ccol!24}81 & \cellcolor{ccol!24}81 & \cellcolor{ccol!24}79 \\
			5. & Brachmann-16 & \cellcolor{ccol!13}42 & \cellcolor{ccol!29}98 & \cellcolor{ccol!21}70 & \cellcolor{ccol!26}88 & \cellcolor{ccol!19}64 & \cellcolor{ccol!23}78 & \cellcolor{ccol!25}84 & \cellcolor{ccol!9}29 & \cellcolor{ccol!2}8 & \cellcolor{ccol!3}10 & \cellcolor{ccol!6}21 & \cellcolor{ccol!1}4 & \cellcolor{ccol!14}46 & \cellcolor{ccol!6}19 & \cellcolor{ccol!16}52 & \cellcolor{ccol!7}22 & \cellcolor{ccol!4}12 & \cellcolor{ccol!2}7 & \cellcolor{ccol!1}3 & \cellcolor{ccol!1}3 & \cellcolor{ccol!0}0 & \cellcolor{ccol!0}0 & \cellcolor{ccol!0}0 & \cellcolor{ccol!2}5 & \cellcolor{ccol!1}3 & \cellcolor{ccol!16}54 \\
			6. & Hodan-15-nr & \cellcolor{ccol!30}100 & \cellcolor{ccol!30}100 & \cellcolor{ccol!28}92 & \cellcolor{ccol!19}62 & \cellcolor{ccol!18}60 & \cellcolor{ccol!28}94 & \cellcolor{ccol!28}93 & \cellcolor{ccol!18}59 & \cellcolor{ccol!19}64 & \cellcolor{ccol!20}67 & \cellcolor{ccol!21}71 & \cellcolor{ccol!22}73 & \cellcolor{ccol!19}62 & \cellcolor{ccol!17}57 & \cellcolor{ccol!15}49 & \cellcolor{ccol!17}56 & \cellcolor{ccol!25}85 & \cellcolor{ccol!21}70 & \cellcolor{ccol!17}57 & \cellcolor{ccol!16}55 & \cellcolor{ccol!18}60 & \cellcolor{ccol!7}23 & \cellcolor{ccol!18}60 & \cellcolor{ccol!25}82 & \cellcolor{ccol!24}81 & \cellcolor{ccol!23}77 \\
			7. & Buch-17-ppfh & \cellcolor{ccol!26}88 & \cellcolor{ccol!30}100 & \cellcolor{ccol!28}94 & \cellcolor{ccol!30}100 & \cellcolor{ccol!30}100 & \cellcolor{ccol!26}88 & \cellcolor{ccol!30}100 & \cellcolor{ccol!15}50 & \cellcolor{ccol!0}1 & \cellcolor{ccol!2}7 & \cellcolor{ccol!0}0 & \cellcolor{ccol!2}5 & \cellcolor{ccol!7}25 & \cellcolor{ccol!5}16 & \cellcolor{ccol!1}4 & \cellcolor{ccol!11}35 & \cellcolor{ccol!11}37 & \cellcolor{ccol!15}48 & \cellcolor{ccol!1}4 & \cellcolor{ccol!3}10 & \cellcolor{ccol!1}4 & \cellcolor{ccol!0}0 & \cellcolor{ccol!0}0 & \cellcolor{ccol!4}12 & \cellcolor{ccol!10}34 & \cellcolor{ccol!15}49 \\
			8. & Kehl-16 & \cellcolor{ccol!7}22 & \cellcolor{ccol!30}100 & \cellcolor{ccol!21}70 & \cellcolor{ccol!22}72 & \cellcolor{ccol!29}96 & \cellcolor{ccol!9}30 & \cellcolor{ccol!21}71 & \cellcolor{ccol!5}17 & \cellcolor{ccol!2}7 & \cellcolor{ccol!3}10 & \cellcolor{ccol!6}18 & \cellcolor{ccol!7}24 & \cellcolor{ccol!7}23 & \cellcolor{ccol!3}10 & \cellcolor{ccol!0}0 & \cellcolor{ccol!1}2 & \cellcolor{ccol!3}11 & \cellcolor{ccol!5}17 & \cellcolor{ccol!1}5 & \cellcolor{ccol!0}1 & \cellcolor{ccol!0}0 & \cellcolor{ccol!3}9 & \cellcolor{ccol!3}12 & \cellcolor{ccol!17}56 & \cellcolor{ccol!15}52 & \cellcolor{ccol!7}22 \\
			9. & Buch-17-si & \cellcolor{ccol!19}62 & \cellcolor{ccol!30}100 & \cellcolor{ccol!28}94 & \cellcolor{ccol!19}62 & \cellcolor{ccol!16}52 & \cellcolor{ccol!10}34 & \cellcolor{ccol!29}97 & \cellcolor{ccol!6}21 & \cellcolor{ccol!0}0 & \cellcolor{ccol!0}1 & \cellcolor{ccol!5}17 & \cellcolor{ccol!5}17 & \cellcolor{ccol!3}9 & \cellcolor{ccol!1}3 & \cellcolor{ccol!0}1 & \cellcolor{ccol!1}4 & \cellcolor{ccol!0}0 & \cellcolor{ccol!2}8 & \cellcolor{ccol!1}2 & \cellcolor{ccol!0}0 & \cellcolor{ccol!0}0 & \cellcolor{ccol!0}0 & \cellcolor{ccol!0}0 & \cellcolor{ccol!6}20 & \cellcolor{ccol!8}26 & \cellcolor{ccol!4}12 \\
			10. & Brachmann-14 & \cellcolor{ccol!29}96 & \cellcolor{ccol!30}100 & \cellcolor{ccol!20}66 & \cellcolor{ccol!22}72 & \cellcolor{ccol!14}46 & \cellcolor{ccol!28}92 & \cellcolor{ccol!8}28 & \cellcolor{ccol!6}20 & \cellcolor{ccol!0}0 & \cellcolor{ccol!0}0 & \cellcolor{ccol!0}1 & \cellcolor{ccol!0}0 & \cellcolor{ccol!0}0 & \cellcolor{ccol!0}0 & \cellcolor{ccol!0}0 & \cellcolor{ccol!0}0 & \cellcolor{ccol!0}1 & \cellcolor{ccol!0}0 & \cellcolor{ccol!0}0 & \cellcolor{ccol!0}0 & \cellcolor{ccol!0}0 & \cellcolor{ccol!0}0 & \cellcolor{ccol!0}0 & \cellcolor{ccol!0}0 & \cellcolor{ccol!0}1 & \cellcolor{ccol!1}2 \\
			11. & Buch-17-ecsad & \cellcolor{ccol!20}66 & \cellcolor{ccol!26}88 & \cellcolor{ccol!0}0 & \cellcolor{ccol!17}56 & \cellcolor{ccol!10}34 & \cellcolor{ccol!0}0 & \cellcolor{ccol!29}95 & \cellcolor{ccol!7}23 & \cellcolor{ccol!0}0 & \cellcolor{ccol!0}0 & \cellcolor{ccol!0}0 & \cellcolor{ccol!0}0 & \cellcolor{ccol!0}0 & \cellcolor{ccol!0}1 & \cellcolor{ccol!0}1 & \cellcolor{ccol!0}0 & \cellcolor{ccol!0}0 & \cellcolor{ccol!0}0 & \cellcolor{ccol!0}0 & \cellcolor{ccol!0}0 & \cellcolor{ccol!0}0 & \cellcolor{ccol!0}0 & \cellcolor{ccol!0}0 & \cellcolor{ccol!0}1 & \cellcolor{ccol!0}0 & \cellcolor{ccol!2}8 \\
			12. & Buch-17-shot & \cellcolor{ccol!16}52 & \cellcolor{ccol!26}88 & \cellcolor{ccol!11}38 & \cellcolor{ccol!11}36 & \cellcolor{ccol!12}40 & \cellcolor{ccol!1}4 & \cellcolor{ccol!20}66 & \cellcolor{ccol!3}11 & \cellcolor{ccol!0}0 & \cellcolor{ccol!0}0 & \cellcolor{ccol!0}1 & \cellcolor{ccol!0}0 & \cellcolor{ccol!0}1 & \cellcolor{ccol!2}5 & \cellcolor{ccol!0}0 & \cellcolor{ccol!1}2 & \cellcolor{ccol!0}1 & \cellcolor{ccol!0}0 & \cellcolor{ccol!0}0 & \cellcolor{ccol!0}1 & \cellcolor{ccol!0}0 & \cellcolor{ccol!0}1 & \cellcolor{ccol!0}1 & \cellcolor{ccol!1}2 & \cellcolor{ccol!0}1 & \cellcolor{ccol!1}3 \\
			13. & Tejani-14 & \cellcolor{ccol!13}42 & \cellcolor{ccol!11}36 & \cellcolor{ccol!0}0 & \cellcolor{ccol!12}40 & \cellcolor{ccol!8}26 & \cellcolor{ccol!22}74 & \cellcolor{ccol!1}4 & \cellcolor{ccol!5}16 & \cellcolor{ccol!0}0 & \cellcolor{ccol!0}1 & \cellcolor{ccol!0}1 & \cellcolor{ccol!0}0 & \cellcolor{ccol!0}0 & \cellcolor{ccol!0}0 & \cellcolor{ccol!0}0 & \cellcolor{ccol!0}0 & \cellcolor{ccol!0}0 & \cellcolor{ccol!0}0 & \cellcolor{ccol!0}0 & \cellcolor{ccol!0}0 & \cellcolor{ccol!0}0 & \cellcolor{ccol!0}0 & \cellcolor{ccol!0}0 & \cellcolor{ccol!0}0 & \cellcolor{ccol!0}0 & \cellcolor{ccol!0}0 \\
			14. & Buch-16-ppfh & \cellcolor{ccol!8}28 & \cellcolor{ccol!10}34 & \cellcolor{ccol!6}20 & \cellcolor{ccol!2}6 & \cellcolor{ccol!7}24 & \cellcolor{ccol!2}8 & \cellcolor{ccol!1}4 & \cellcolor{ccol!0}1 & \cellcolor{ccol!0}1 & \cellcolor{ccol!2}6 & \cellcolor{ccol!1}3 & \cellcolor{ccol!0}1 & \cellcolor{ccol!7}24 & \cellcolor{ccol!1}4 & \cellcolor{ccol!3}10 & \cellcolor{ccol!4}13 & \cellcolor{ccol!3}10 & \cellcolor{ccol!4}13 & \cellcolor{ccol!1}3 & \cellcolor{ccol!2}8 & \cellcolor{ccol!0}1 & \cellcolor{ccol!0}0 & \cellcolor{ccol!0}0 & \cellcolor{ccol!1}5 & \cellcolor{ccol!10}32 & \cellcolor{ccol!4}13 \\
			15. & Buch-16-ecsad & \cellcolor{ccol!1}4 & \cellcolor{ccol!1}4 & \cellcolor{ccol!2}8 & \cellcolor{ccol!1}4 & \cellcolor{ccol!1}2 & \cellcolor{ccol!0}0 & \cellcolor{ccol!2}5 & \cellcolor{ccol!1}3 & \cellcolor{ccol!0}0 & \cellcolor{ccol!0}1 & \cellcolor{ccol!0}0 & \cellcolor{ccol!0}0 & \cellcolor{ccol!0}0 & \cellcolor{ccol!0}0 & \cellcolor{ccol!0}0 & \cellcolor{ccol!0}1 & \cellcolor{ccol!0}0 & \cellcolor{ccol!0}0 & \cellcolor{ccol!0}0 & \cellcolor{ccol!0}0 & \cellcolor{ccol!0}0 & \cellcolor{ccol!0}0 & \cellcolor{ccol!0}0 & \cellcolor{ccol!0}0 & \cellcolor{ccol!0}0 & \cellcolor{ccol!1}2 \\

			\toprule
			
			& {\tiny } & \multicolumn{12}{c}{{\tiny T-LESS}} & \multicolumn{14}{c}{{\tiny RU-APC}} \\
			\cmidrule(l{2pt}r{2pt}){3-14} \cmidrule(l{2pt}r{2pt}){15-28}
			& & 19 & 20 & 21 & 22 & 23 & 24 & 25 & 26 & 27 & 28 & 29 & 30 & 1 & 2 & 3 & 4 & 5 & 6 & 7 & 8 & 9 & 10 & 11 & 12 & 13 & 14 \\
			\midrule
			
			1. & Vidal-18 & \cellcolor{ccol!17}57 & \cellcolor{ccol!13}43 & \cellcolor{ccol!19}62 & \cellcolor{ccol!21}69 & \cellcolor{ccol!26}85 & \cellcolor{ccol!20}66 & \cellcolor{ccol!13}43 & \cellcolor{ccol!17}58 & \cellcolor{ccol!19}62 & \cellcolor{ccol!21}69 & \cellcolor{ccol!21}69 & \cellcolor{ccol!25}85 & \cellcolor{ccol!12}39 & \cellcolor{ccol!11}38 & \cellcolor{ccol!13}42 & \cellcolor{ccol!16}54 & \cellcolor{ccol!16}53 & \cellcolor{ccol!13}43 & \cellcolor{ccol!1}4 & \cellcolor{ccol!25}82 & \cellcolor{ccol!10}32 & \cellcolor{ccol!0}0 & \cellcolor{ccol!14}48 & \cellcolor{ccol!14}47 & \cellcolor{ccol!6}20 & \cellcolor{ccol!2}8 \\
			2. & Drost-10-edge & \cellcolor{ccol!17}55 & \cellcolor{ccol!14}47 & \cellcolor{ccol!16}55 & \cellcolor{ccol!17}56 & \cellcolor{ccol!25}84 & \cellcolor{ccol!18}59 & \cellcolor{ccol!14}47 & \cellcolor{ccol!21}69 & \cellcolor{ccol!18}61 & \cellcolor{ccol!24}80 & \cellcolor{ccol!25}84 & \cellcolor{ccol!27}89 & \cellcolor{ccol!0}0 & \cellcolor{ccol!6}20 & \cellcolor{ccol!11}35 & \cellcolor{ccol!14}47 & \cellcolor{ccol!11}35 & \cellcolor{ccol!12}39 & \cellcolor{ccol!0}0 & \cellcolor{ccol!27}89 & \cellcolor{ccol!8}28 & \cellcolor{ccol!0}0 & \cellcolor{ccol!14}48 & \cellcolor{ccol!6}21 & \cellcolor{ccol!4}15 & \cellcolor{ccol!1}3 \\
			3. & Drost-10 & \cellcolor{ccol!16}53 & \cellcolor{ccol!10}35 & \cellcolor{ccol!18}60 & \cellcolor{ccol!18}61 & \cellcolor{ccol!24}81 & \cellcolor{ccol!17}57 & \cellcolor{ccol!8}28 & \cellcolor{ccol!15}51 & \cellcolor{ccol!10}32 & \cellcolor{ccol!18}60 & \cellcolor{ccol!24}81 & \cellcolor{ccol!21}71 & \cellcolor{ccol!0}0 & \cellcolor{ccol!3}11 & \cellcolor{ccol!9}29 & \cellcolor{ccol!14}45 & \cellcolor{ccol!10}33 & \cellcolor{ccol!9}29 & \cellcolor{ccol!8}26 & \cellcolor{ccol!21}71 & \cellcolor{ccol!3}10 & \cellcolor{ccol!0}0 & \cellcolor{ccol!14}47 & \cellcolor{ccol!3}9 & \cellcolor{ccol!0}0 & \cellcolor{ccol!0}0 \\
			4. & Hodan-15 & \cellcolor{ccol!18}59 & \cellcolor{ccol!8}27 & \cellcolor{ccol!17}57 & \cellcolor{ccol!15}50 & \cellcolor{ccol!22}74 & \cellcolor{ccol!18}59 & \cellcolor{ccol!14}47 & \cellcolor{ccol!22}72 & \cellcolor{ccol!13}45 & \cellcolor{ccol!22}73 & \cellcolor{ccol!22}74 & \cellcolor{ccol!26}85 & \cellcolor{ccol!1}4 & \cellcolor{ccol!11}36 & \cellcolor{ccol!18}59 & \cellcolor{ccol!7}24 & \cellcolor{ccol!14}47 & \cellcolor{ccol!14}46 & \cellcolor{ccol!16}52 & \cellcolor{ccol!29}97 & \cellcolor{ccol!8}28 & \cellcolor{ccol!8}28 & \cellcolor{ccol!10}34 & \cellcolor{ccol!16}52 & \cellcolor{ccol!5}17 & \cellcolor{ccol!0}0 \\
			5. & Brachmann-16 & \cellcolor{ccol!11}38 & \cellcolor{ccol!0}1 & \cellcolor{ccol!12}39 & \cellcolor{ccol!6}19 & \cellcolor{ccol!18}61 & \cellcolor{ccol!0}1 & \cellcolor{ccol!5}16 & \cellcolor{ccol!8}27 & \cellcolor{ccol!5}17 & \cellcolor{ccol!4}13 & \cellcolor{ccol!2}6 & \cellcolor{ccol!2}5 & \cellcolor{ccol!2}6 & \cellcolor{ccol!19}64 & \cellcolor{ccol!8}25 & \cellcolor{ccol!6}21 & \cellcolor{ccol!10}32 & \cellcolor{ccol!12}41 & \cellcolor{ccol!14}47 & \cellcolor{ccol!11}37 & \cellcolor{ccol!0}1 & \cellcolor{ccol!0}0 & \cellcolor{ccol!6}18 & \cellcolor{ccol!12}40 & \cellcolor{ccol!0}0 & \cellcolor{ccol!2}5 \\
			6. & Hodan-15-nr & \cellcolor{ccol!17}58 & \cellcolor{ccol!8}27 & \cellcolor{ccol!17}55 & \cellcolor{ccol!15}50 & \cellcolor{ccol!22}73 & \cellcolor{ccol!18}60 & \cellcolor{ccol!15}49 & \cellcolor{ccol!22}72 & \cellcolor{ccol!12}40 & \cellcolor{ccol!22}72 & \cellcolor{ccol!23}76 & \cellcolor{ccol!26}85 & \cellcolor{ccol!1}4 & \cellcolor{ccol!12}39 & \cellcolor{ccol!15}50 & \cellcolor{ccol!7}24 & \cellcolor{ccol!12}41 & \cellcolor{ccol!5}15 & \cellcolor{ccol!13}43 & \cellcolor{ccol!27}91 & \cellcolor{ccol!8}25 & \cellcolor{ccol!10}33 & \cellcolor{ccol!9}31 & \cellcolor{ccol!12}39 & \cellcolor{ccol!5}16 & \cellcolor{ccol!0}1 \\
			7. & Buch-17-ppfh & \cellcolor{ccol!9}31 & \cellcolor{ccol!8}25 & \cellcolor{ccol!11}36 & \cellcolor{ccol!10}35 & \cellcolor{ccol!21}71 & \cellcolor{ccol!14}46 & \cellcolor{ccol!19}64 & \cellcolor{ccol!15}51 & \cellcolor{ccol!1}4 & \cellcolor{ccol!13}44 & \cellcolor{ccol!15}49 & \cellcolor{ccol!17}58 & \cellcolor{ccol!5}16 & \cellcolor{ccol!2}5 & \cellcolor{ccol!5}17 & \cellcolor{ccol!15}51 & \cellcolor{ccol!8}27 & \cellcolor{ccol!2}6 & \cellcolor{ccol!17}57 & \cellcolor{ccol!7}24 & \cellcolor{ccol!3}8 & \cellcolor{ccol!3}10 & \cellcolor{ccol!17}55 & \cellcolor{ccol!2}5 & \cellcolor{ccol!3}11 & \cellcolor{ccol!0}0 \\
			8. & Kehl-16 & \cellcolor{ccol!11}35 & \cellcolor{ccol!1}5 & \cellcolor{ccol!8}26 & \cellcolor{ccol!8}27 & \cellcolor{ccol!21}71 & \cellcolor{ccol!11}36 & \cellcolor{ccol!8}28 & \cellcolor{ccol!15}51 & \cellcolor{ccol!10}34 & \cellcolor{ccol!16}54 & \cellcolor{ccol!26}86 & \cellcolor{ccol!21}69 & \cellcolor{ccol!6}19 & \cellcolor{ccol!4}14 & \cellcolor{ccol!14}46 & \cellcolor{ccol!11}38 & \cellcolor{ccol!16}54 & \cellcolor{ccol!12}40 & \cellcolor{ccol!1}4 & \cellcolor{ccol!24}80 & \cellcolor{ccol!1}3 & \cellcolor{ccol!2}5 & \cellcolor{ccol!1}3 & \cellcolor{ccol!11}37 & \cellcolor{ccol!2}7 & \cellcolor{ccol!2}5 \\
			9. & Buch-17-si & \cellcolor{ccol!3}11 & \cellcolor{ccol!6}21 & \cellcolor{ccol!5}18 & \cellcolor{ccol!3}11 & \cellcolor{ccol!11}37 & \cellcolor{ccol!1}4 & \cellcolor{ccol!15}52 & \cellcolor{ccol!16}53 & \cellcolor{ccol!1}3 & \cellcolor{ccol!10}35 & \cellcolor{ccol!9}32 & \cellcolor{ccol!16}53 & \cellcolor{ccol!7}24 & \cellcolor{ccol!15}49 & \cellcolor{ccol!5}16 & \cellcolor{ccol!12}39 & \cellcolor{ccol!1}3 & \cellcolor{ccol!1}4 & \cellcolor{ccol!10}32 & \cellcolor{ccol!16}54 & \cellcolor{ccol!4}14 & \cellcolor{ccol!3}9 & \cellcolor{ccol!13}43 & \cellcolor{ccol!5}15 & \cellcolor{ccol!5}17 & \cellcolor{ccol!2}5 \\
			10. & Brachmann-14 & \cellcolor{ccol!0}0 & \cellcolor{ccol!0}0 & \cellcolor{ccol!0}0 & \cellcolor{ccol!0}0 & \cellcolor{ccol!0}1 & \cellcolor{ccol!0}0 & \cellcolor{ccol!0}1 & \cellcolor{ccol!0}1 & \cellcolor{ccol!0}0 & \cellcolor{ccol!0}0 & \cellcolor{ccol!0}0 & \cellcolor{ccol!0}0 & \cellcolor{ccol!2}6 & \cellcolor{ccol!24}80 & \cellcolor{ccol!13}42 & \cellcolor{ccol!6}19 & \cellcolor{ccol!9}31 & \cellcolor{ccol!10}33 & \cellcolor{ccol!16}52 & \cellcolor{ccol!27}89 & \cellcolor{ccol!6}19 & \cellcolor{ccol!0}1 & \cellcolor{ccol!0}0 & \cellcolor{ccol!12}40 & \cellcolor{ccol!2}7 & \cellcolor{ccol!0}0 \\
			11. & Buch-17-ecsad & \cellcolor{ccol!5}16 & \cellcolor{ccol!3}11 & \cellcolor{ccol!5}16 & \cellcolor{ccol!2}8 & \cellcolor{ccol!8}27 & \cellcolor{ccol!6}20 & \cellcolor{ccol!15}51 & \cellcolor{ccol!9}31 & \cellcolor{ccol!0}0 & \cellcolor{ccol!10}32 & \cellcolor{ccol!7}22 & \cellcolor{ccol!1}3 & \cellcolor{ccol!0}1 & \cellcolor{ccol!1}2 & \cellcolor{ccol!0}0 & \cellcolor{ccol!0}1 & \cellcolor{ccol!1}3 & \cellcolor{ccol!2}8 & \cellcolor{ccol!7}23 & \cellcolor{ccol!10}34 & \cellcolor{ccol!2}5 & \cellcolor{ccol!2}8 & \cellcolor{ccol!1}2 & \cellcolor{ccol!0}0 & \cellcolor{ccol!1}3 & \cellcolor{ccol!0}1 \\
			12. & Buch-17-shot & \cellcolor{ccol!2}6 & \cellcolor{ccol!2}6 & \cellcolor{ccol!2}8 & \cellcolor{ccol!1}2 & \cellcolor{ccol!9}28 & \cellcolor{ccol!1}3 & \cellcolor{ccol!5}17 & \cellcolor{ccol!4}13 & \cellcolor{ccol!0}0 & \cellcolor{ccol!3}11 & \cellcolor{ccol!2}7 & \cellcolor{ccol!2}6 & \cellcolor{ccol!0}0 & \cellcolor{ccol!0}0 & \cellcolor{ccol!0}0 & \cellcolor{ccol!0}0 & \cellcolor{ccol!0}0 & \cellcolor{ccol!0}0 & \cellcolor{ccol!0}0 & \cellcolor{ccol!0}1 & \cellcolor{ccol!0}0 & \cellcolor{ccol!0}0 & \cellcolor{ccol!0}0 & \cellcolor{ccol!0}0 & \cellcolor{ccol!0}0 & \cellcolor{ccol!0}0 \\
			13. & Tejani-14 & \cellcolor{ccol!0}0 & \cellcolor{ccol!0}0 & \cellcolor{ccol!0}0 & \cellcolor{ccol!0}0 & \cellcolor{ccol!0}0 & \cellcolor{ccol!0}0 & \cellcolor{ccol!0}0 & \cellcolor{ccol!0}0 & \cellcolor{ccol!0}0 & \cellcolor{ccol!0}0 & \cellcolor{ccol!0}0 & \cellcolor{ccol!1}2 & \cellcolor{ccol!0}1 & \cellcolor{ccol!0}0 & \cellcolor{ccol!0}0 & \cellcolor{ccol!1}3 & \cellcolor{ccol!3}9 & \cellcolor{ccol!0}0 & \cellcolor{ccol!0}0 & \cellcolor{ccol!2}5 & \cellcolor{ccol!0}0 & \cellcolor{ccol!0}0 & \cellcolor{ccol!0}0 & \cellcolor{ccol!1}3 & \cellcolor{ccol!0}0 & \cellcolor{ccol!0}0 \\
			14. & Buch-16-ppfh & \cellcolor{ccol!1}3 & \cellcolor{ccol!1}3 & \cellcolor{ccol!3}8 & \cellcolor{ccol!2}8 & \cellcolor{ccol!5}16 & \cellcolor{ccol!1}2 & \cellcolor{ccol!7}24 & \cellcolor{ccol!1}4 & \cellcolor{ccol!2}5 & \cellcolor{ccol!3}11 & \cellcolor{ccol!2}6 & \cellcolor{ccol!0}1 & \cellcolor{ccol!0}0 & \cellcolor{ccol!0}0 & \cellcolor{ccol!2}6 & \cellcolor{ccol!6}19 & \cellcolor{ccol!1}2 & \cellcolor{ccol!4}12 & \cellcolor{ccol!10}34 & \cellcolor{ccol!2}8 & \cellcolor{ccol!0}0 & \cellcolor{ccol!0}0 & \cellcolor{ccol!11}38 & \cellcolor{ccol!1}2 & \cellcolor{ccol!2}5 & \cellcolor{ccol!0}0 \\
			15. & Buch-16-ecsad & \cellcolor{ccol!1}2 & \cellcolor{ccol!0}1 & \cellcolor{ccol!1}3 & \cellcolor{ccol!0}0 & \cellcolor{ccol!3}10 & \cellcolor{ccol!0}0 & \cellcolor{ccol!4}12 & \cellcolor{ccol!0}1 & \cellcolor{ccol!0}2 & \cellcolor{ccol!1}4 & \cellcolor{ccol!0}1 & \cellcolor{ccol!0}1 & \cellcolor{ccol!0}0 & \cellcolor{ccol!1}3 & \cellcolor{ccol!2}5 & \cellcolor{ccol!0}0 & \cellcolor{ccol!0}1 & \cellcolor{ccol!0}1 & \cellcolor{ccol!3}11 & \cellcolor{ccol!4}13 & \cellcolor{ccol!0}0 & \cellcolor{ccol!0}0 & \cellcolor{ccol!1}3 & \cellcolor{ccol!1}2 & \cellcolor{ccol!0}0 & \cellcolor{ccol!0}1 \\

			\bottomrule
		\end{tabularx}
		
		\captionof{table}{\label{tab:eval_per_object} Recall scores (\%) per object for $\tau=20\,\si{mm}$ and $\theta=0.3$.}
	\end{center}
\end{figure}

\begin{figure}[!h]
	\begin{center}
		\scriptsize
		\begin{tabularx}{\textwidth}{ c c c }
			\includegraphics[width=0.325\columnwidth]{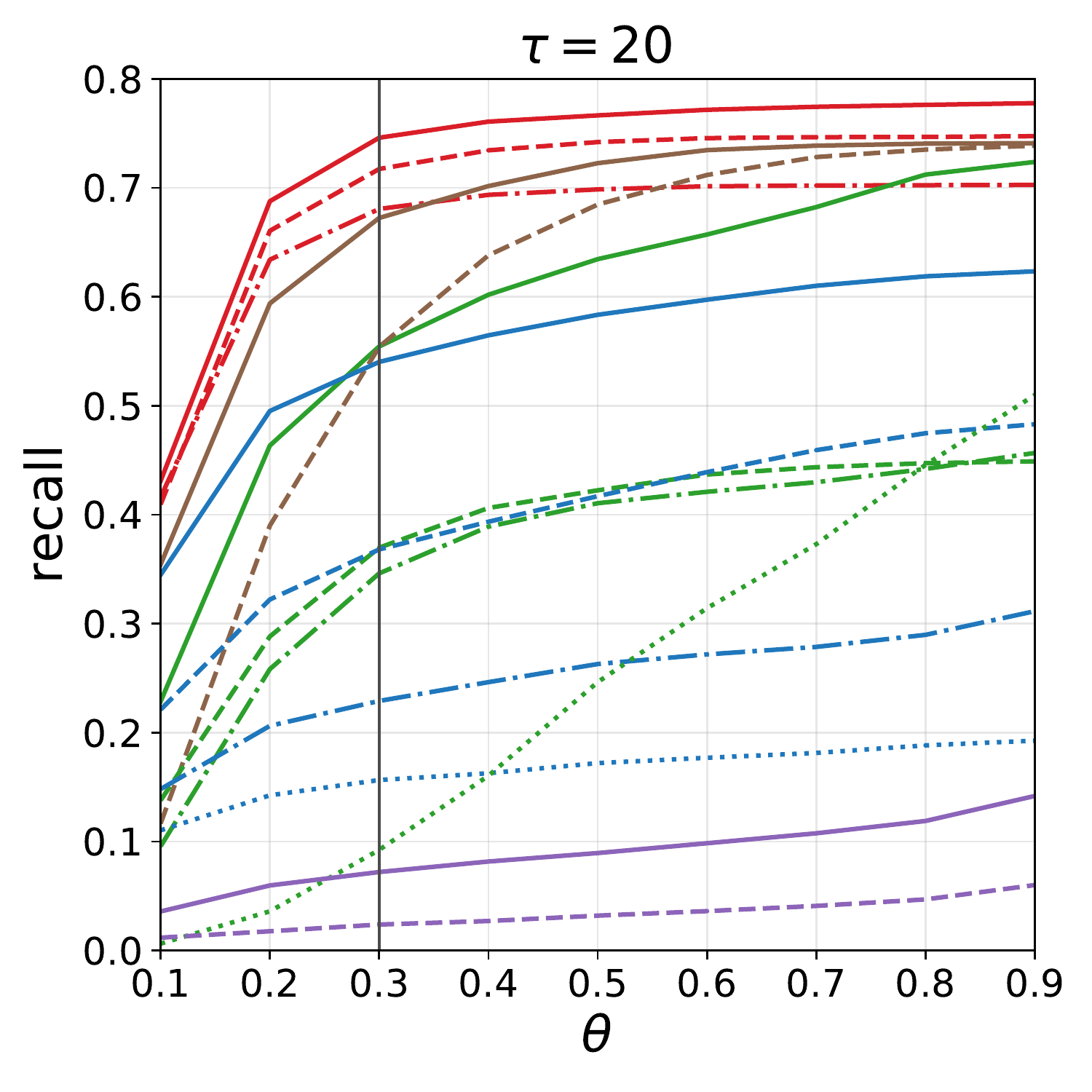} &
			\includegraphics[width=0.325\columnwidth]{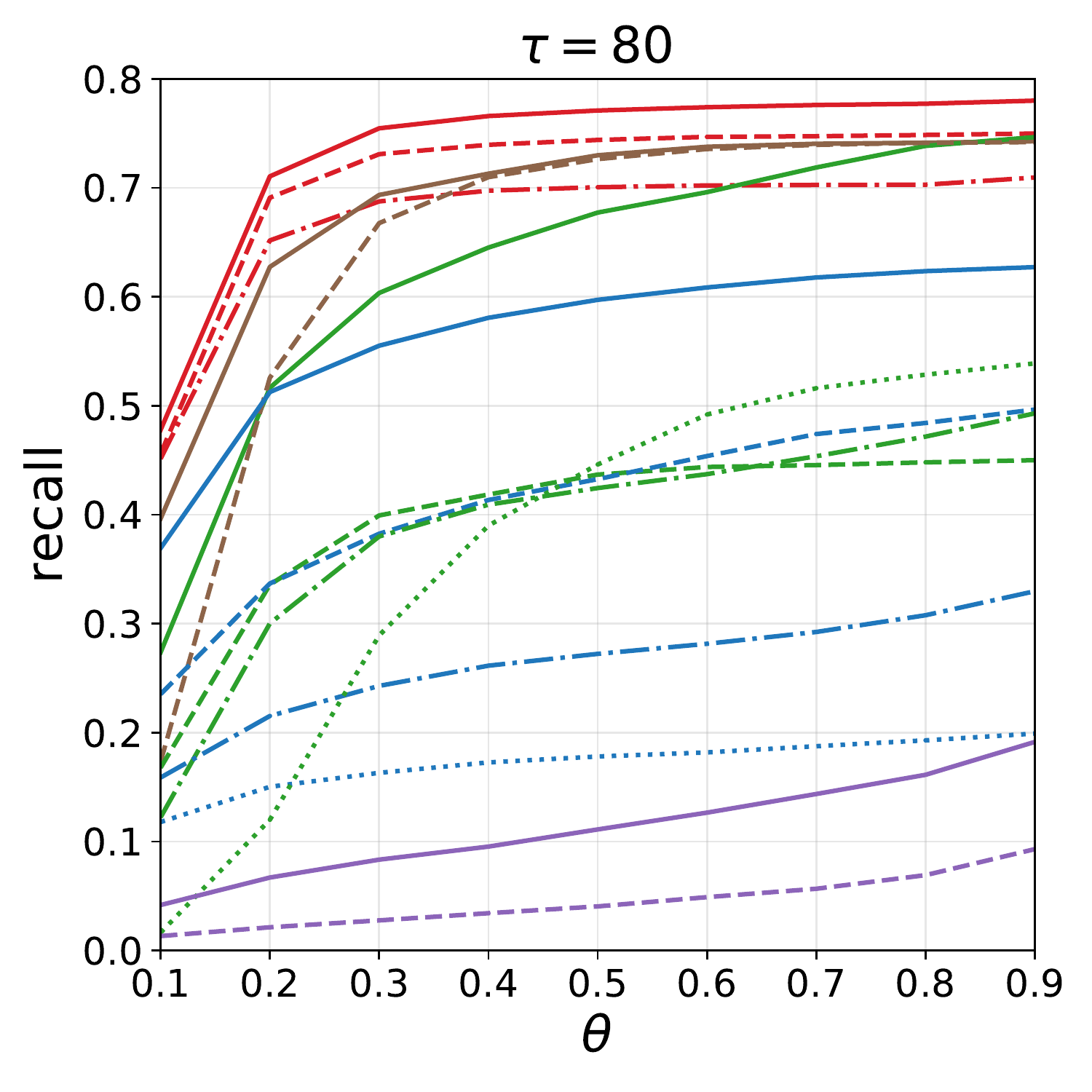} &
			\includegraphics[width=0.325\columnwidth]{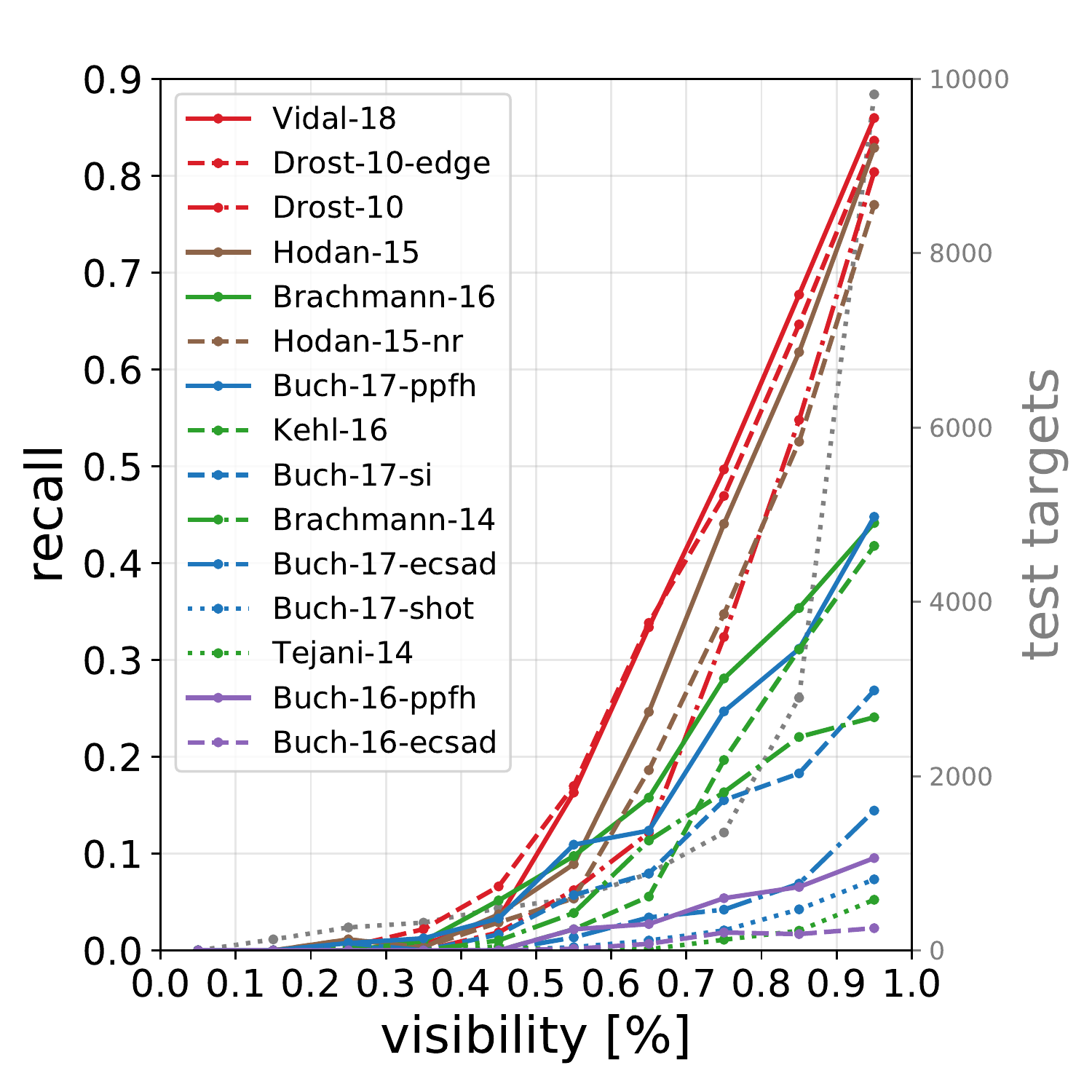} \\
		\end{tabularx}
		\vspace{-1ex}		
		\caption{\label{tab:eval_curves} Left, middle: Average of the per-dataset recall scores for the misalignment tolerance~$\tau$ fixed to $20\,\si{mm}$ and $80\,\si{mm}$, and varying value of the correctness threshold~$\theta$. The curves do not change much for $\tau > 80\,\si{mm}$. Right: The recall scores w.r.t. the visible fraction of the target object. If more instances of the target object were present in the test image, the largest visible fraction was considered.}
		\vspace{1ex}	
	\end{center}
\end{figure}

\subsection{Results}

\subsubsection{Accuracy.}

Tab.~\ref{tab:eval_per_dataset} and \ref{tab:eval_per_object} show the recall scores of the evaluated methods per dataset and per object respectively, for the misalignment tolerance $\tau=20\,\si{mm}$ and the correctness threshold $\theta=0.3$. Ranking of the methods according to the recall score is mostly stable across the datasets. Methods based on point-pair features perform best. Vidal-18 is the top-performing method with the average recall of 74.6\%, followed by Drost-10-edge, Drost-10, and the template matching method Hodaň-15, all with the average recall above 67\%. Brachmann-16 is the best learning-based method, with 55.4\%, and Buch-17-ppfh is the best method based on 3D local features, with 54.0\%. Scores of Buch-16-si and Buch-16-shot are inferior to the other variants of this method and not presented.

Fig.~\ref{tab:eval_curves} shows the average of the per-dataset recall scores for different values of $\tau$ and $\theta$. If the misalignment tolerance $\tau$ is increased from $20\,\si{mm}$ to $80\,\si{mm}$, the scores increase only slightly for most methods. Similarly, the scores increase only slowly for $\theta > 0.3$. This suggests that poses estimated by most methods are either of a high quality or totally off, \ie it is a hit or miss.

\subsubsection{Speed.}

The average running times per test target are reported in Tab.~\ref{tab:eval_per_dataset}. However, the methods were evaluated on different computers\footnote{Specifications of computers used for the evaluation are on the project website.} and thus the presented running times are not directly comparable.
Moreover, the methods were optimized primarily for the recall score, not for speed.
For example, we evaluated Drost-10 with several parameter settings and observed that the running time can be lowered by a factor of $\mytilde5$ to $0.5\,\si{s}$ with only a relatively small drop of the average recall score from $68.1\%$ to $65.8\%$. However, in Tab.~\ref{tab:eval_per_dataset} we present the result with the highest score. Brachmann-14 could be sped up by sub-sampling the 3D object models and Hodaň-15 by using less object templates. A study of such speed/accuracy trade-offs is left for future work.

\subsubsection{Open Problems.} Occlusion is a big challenge for current methods, as shown by scores dropping swiftly already at low levels of occlusion (Fig.~\ref{tab:eval_curves}, right). The big gap between LM and LM-O scores provide further evidence. All methods perform on LM by at least 30\% better than on LM-O, which includes the same but partially occluded objects. Inspection of estimated poses on T-LESS test images confirms the weak performance for occluded objects.
Scores on TUD-L show that varying lighting conditions present a serious challenge for methods that rely on synthetic training RGB images, which were generated with fixed lighting. Methods relying only on depth information (\eg Vidal-18, Drost-10) are noticeably more robust under such conditions.
Note that Brach\-mann-16 achieved a high score on TUD-L despite relying on RGB images because it used real training images, which were captured under the same range of lighting conditions as the test images.
Methods based on 3D local features and learning-based methods have very low scores
on T-LESS, which is likely caused by the object symmetries and similarities.
All methods perform poorly on RU-APC, which is likely because of a higher level of noise in the depth images.

\section{Conclusion}

We have proposed a benchmark for 6D object pose estimation that includes eight datasets in a unified format, an evaluation methodology, a comprehensive evaluation of 15 recent methods, and an online evaluation system open for continuous submission of new results. With this benchmark, we have captured the status quo in the field and will be able to systematically measure its progress in the future. The evaluation showed that methods based on point-pair features perform best, outperforming template matching methods, learning-based methods and methods based on 3D local features.
As open problems, our analysis identified occlusion, varying lighting conditions, and object symmetries and similarities.

\section*{Acknowledgements}

\noindent We gratefully acknowledge Manolis Lourakis, Joachim Staib, Christoph Kick, 
Juil Sock and Pavel Haluza for their help. This work was supported by CTU student grant SGS17/185/OHK3/3T/13, Technology Agency of the Czech Republic research program TE01020415 (V3C -- Visual Computing Competence Center), and the project for GA\v{C}R, No. 16-072105: Complex network methods applied to ancient Egyptian data in the Old Kingdom (2700–2180 BC).

{\small
\bibliographystyle{splncs04}
\bibliography{ref}
}

\end{document}